\documentclass{article}

\usepackage[nonatbib,preprint]{neurips_2024}
\usepackage[numbers]{natbib}

\usepackage[utf8]{inputenc} 
\usepackage[T1]{fontenc}    
\usepackage{hyperref}       
\usepackage{url}            
\usepackage{booktabs}       
\usepackage{amsfonts}       
\usepackage{nicefrac}       
\usepackage{microtype}      
\usepackage{xcolor}         

\usepackage{multirow}
\usepackage{amsmath}
\usepackage{makecell}
\usepackage{multirow}
\usepackage{mathrsfs}
\usepackage[ruled]{algorithm2e}
\usepackage{cellspace}

\usepackage{makecell}
\usepackage{bm}
\usepackage{graphicx}
\usepackage{paralist}
\usepackage{enumitem}
\usepackage[utf8]{inputenc} 
\usepackage[T1]{fontenc}    
\usepackage{hyperref}       
\usepackage{url}            
\usepackage{booktabs}       
\usepackage{amsfonts}       
\usepackage{nicefrac}       
\usepackage{microtype}      
\usepackage[ruled]{algorithm2e}
\usepackage{threeparttable}

\usepackage{graphicx}
\usepackage{subfigure}
\usepackage{booktabs} 
\usepackage{caption}

\usepackage{framed}
\usepackage{amssymb}
\usepackage{adjustbox}
\usepackage{amsfonts}
\usepackage{mathrsfs}
\usepackage{mathtools}
\usepackage{array}
\usepackage{amsthm}
\usepackage{verbatim} 
\usepackage{enumerate}
\usepackage{bbm}
\usepackage{commath}
\usepackage{wrapfig}
\usepackage{amsbsy}
\usepackage{float}
\usepackage{amsmath}
\usepackage{tabularx}
\usepackage{listings}
\usepackage{xcolor}
\usepackage{colortbl}
\usepackage{tcolorbox}
\usepackage{bbding}
\usepackage{multirow}
\usepackage[toc,page,header]{appendix}
\usepackage{minitoc}
\usepackage{titletoc}
\usepackage{CJKutf8}

\newcommand{\vpara}[1]{\vspace{0.07in}\noindent\textbf{#1 }}
\newcommand{\model}{MathGLM-Vision\xspace}
\newcommand{\data}{MathVL\xspace}

\title{MathGLM-Vision: Solving Mathematical Problems with Multi-Modal Large Language Model}

\author{%
  Zhen Yang$^{13\dagger*}$,
  Jinhao Chen$^{23\dagger*}$,
  Zhengxiao Du$^{13}$,
  Wenmeng Yu$^{3}$,
  Weihan Wang$^{13}$,\\
  \textbf{Wenyi Hong$^{13}$,
  Zhihuan Jiang$^{13}$,
  Bin Xu$^{1}$,
  Jie Tang$^{1}$ }\\ \\
  \textsuperscript{1}Tsinghua University \quad
  \textsuperscript{2}Beihang University \quad
  \textsuperscript{3}Zhipu.AI 
\\ \\
}

\begin{document}

\maketitle

\renewcommand{\thefootnote}{\fnsymbol{footnote}}
    \footnotetext[1]{ZY and JHC contributed equally. Emails: \texttt{yangz21@mails.tsinghua.edu.cn, chenjh24@buaa.edu.cn}}
    \footnotetext[2]{Work done while Zhen Yang and Jinhao Chen interned at Zhipu AI.}
\renewcommand{\thefootnote}{\arabic{footnote}}

\begin{abstract}

Large language models (LLMs) have demonstrated significant capabilities in mathematical reasoning, particularly with text-based mathematical problems. However, current multi-modal large language models (MLLMs), especially those specialized in mathematics, tend to focus predominantly on solving geometric problems but ignore the diversity of visual information available in other areas of mathematics. Moreover, the geometric information for these specialized mathematical MLLMs is derived from several public datasets, which are typically limited in diversity and complexity. To address these limitations, we aim to construct a fine-tuning dataset named \data, and develop a series of specialized mathematical MLLMs termed \model by conducting Supervised Fine-Tuning (SFT) on \data with various parameter-scale backbones. To extensively evaluate the effectiveness of \model, we conduct experiments on several public benchmarks and our curated \data-test consisting of 2,000 problems. Experimental results demonstrate that \model achieves significant improvements compared with some existing models, including backbone models and open-source mathematical MLLMs. These findings indicate the importance of diversity dataset in enhancing the mathematical reasoning abilities of MLLMs.

\end{abstract}

\section{Introduction}
Recent advancements in computational linguistics have led to substantial progress in solving mathematical problems using Large Language Models (LLMs) with multi-step reasoning processes~\cite{lightman2023let}. For example, models like GPT-4~\cite{achiam2023gpt}, Qwen~\cite{bai2023qwen}, GLM-4~\cite{team2024chatglm}, LLaMA~\cite{touvron2023llama1,touvron2023llama2} have demonstrated impressive performance on mathematical datasets such as GSM8K~\cite{cobbe2021training} and MATH~\cite{hendrycks2021measuring}. Furthermore, the development of specialized mathematical models is expanding the potential of LLMs in this domain. These models, specifically designed for mathematical problem solving, include notable contributions such as WizardMath~\cite{luo2023wizardmath}, MAmmoTH~\cite{yue2023mammoth}, MathCoder~\cite{wang2023mathcoder}, MetaMath~\cite{yu2023metamath}, DeepSeekMath~\cite{shao2024deepseekmath}, and others~\cite{yang2023gpt,yuan2023scaling,gou2023tora,yue2024mammoth2,mitra2024orca,ying2024internlm}. These advancements highlight the growing proficiency of LLMs in handling intricate mathematical reasoning and problem-solving tasks.

Despite significant advancements, the majority of models designed for mathematical problem solving still rely predominately on textual representations. This limits their effectiveness in scenarios that require visual information. Notably, approximately 63\% of mathematics questions in Chinese K12 education include visual elements, highlighting the critical role of visual information in comprehending and solving mathematical problems.


\begin{figure}[thbp]
\centering
\begin{minipage}{.63\textwidth}
  \centering
  \includegraphics[width=\linewidth]{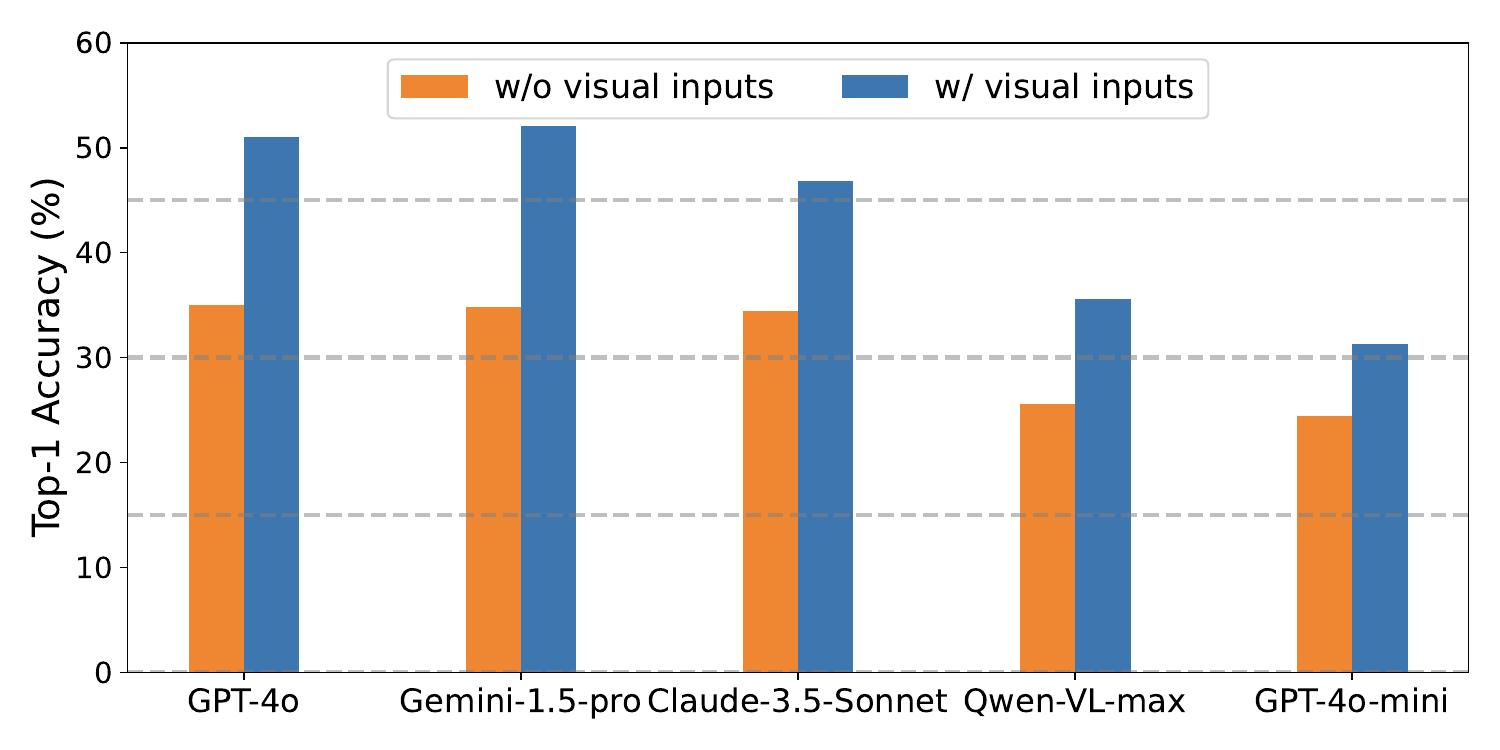}
\end{minipage}%
\begin{minipage}{.36\textwidth}
  \centering
  \includegraphics[width=\linewidth]{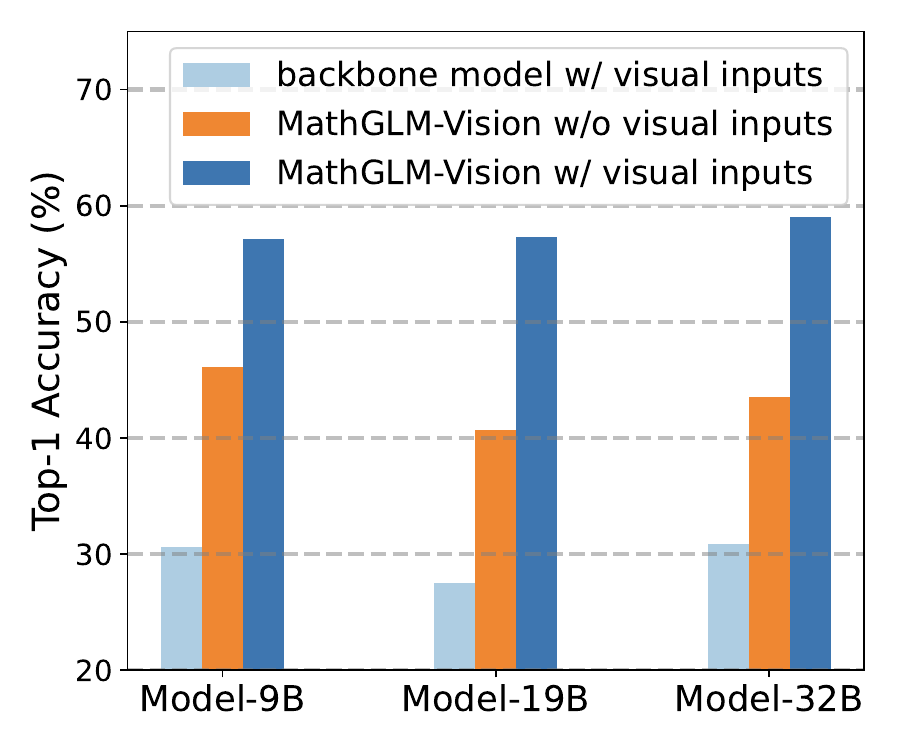}
\end{minipage}
\caption{Insight experiments demonstrates the significance of visual information in solving mathematical problems. (Left) A performance comparison of different models with and without visual inputs. (Right) The accuracies of \model on \data-test with and without visual inputs.}
\label{fig:insight_results}
\end{figure}

Therefore, a crucial question arises: Is visual information essential for solving these mathematical problems that include visual elements? To verify this, we conduct a series of insightful experiments comparing the performance of these models such as GPT-4o, Claude-3.5-Sonnet, Qwen-VL-Max, and Gemini-1.5-Pro on \data-test, both with and without visual inputs. As shown in Figure~\ref{fig:insight_results}, the results clearly demonstrate that the inclusion of visual elements significantly enhances the models' ability to accurately solve complex mathematical problems. Conversely, the exclusion of visual information leads to a pronounced decrease in performance, emphasizing the essential role that visual context plays in solving mathematical problems that incorporate visual elements.





Currently, multi-modal large language models (MLLMs) are at the forefront of efforts to integrate visual and textual information for solving mathematical problems. Close-source models such as GPT-4V~\cite{openai2023gpt4v}, Gemini~\cite{team2023gemini}, Claude3~\cite{Claude3}, Qwen-VL~\cite{bai2023qwen-vl}, along with several open-source MLLMs like CogVLM~\cite{wang2023cogvlm}, MiniGPT~\cite{zhu2023minigpt}, LLaVA-1.5~\cite{liu2024improved}, SPHINX-MoE~\cite{gao2024sphinx}, and LLaVA-NeXT~\cite{liu2024llava}, demonstrate substantial potential in addressing geometric reasoning challenges. Additionally, specialized geometric MLLMs like G-LLaVA~\cite{gao2023g}, GeoGPT4V~\cite{cai2024geogpt4v} and Math-LLaVA~\cite{shihu2024mathllava} are particularly focused on enhancing capabilities in this domain. However, these models still face several challenges and limitations that need to be addressed. 
\begin{itemize}[leftmargin=*]
    \item \textit{Current MLLMs, particularly those specialized in mathematics, predominantly focus on solving geometric problems and tend to overlook the diversity of visual information in mathematics. This visual information encompasses a broad spectrum of elements, including arithmetic, statistics, algebra and word problems, each integral to different mathematical domains beyond geometry.}

    \item \textit{Current fine-tuning dataset for specialized mathematical MLLMs, typically sourced from public datasets like GeoQA and Geometry3K, often lack diversity and complexity. This limitation restricts the models' ability to effectively solve a broader range of mathematical problems.}

    \item \textit{Current specialized mathematical MLLMs are predominantly designed to process single-image inputs and lack the capability to handle multiple images simultaneously. This limitation hampers their ability to tackle complex problems that necessitate the integration of information from multiple visual sources.}
\end{itemize}

In response to these challenges and limitations, we introduce \textbf{\model}, a promising specialized mathematical multi-modal large language model designed to seamlessly integrate visual information with textual analysis. \model is designed to enhance the model's ability to interpret and solve complex mathematical problems involving visual elements, thereby expanding the range of problems that can be addressed effectively. As demonstrated in Figure~\ref{fig:insight_results}, we compare the performance of \model on \data-test with and without visual inputs. The results clearly indicate that \model incorporating visual inputs, significantly outperforms its text-only inputs that ignores visual information. This enhanced capability highlights the importance of integrating multi-modal inputs in advancing the performance of specialized mathematical models.

\begin{figure}[thbp]
\centering
\begin{minipage}{.43\textwidth}
  \centering
  \includegraphics[width=\linewidth]{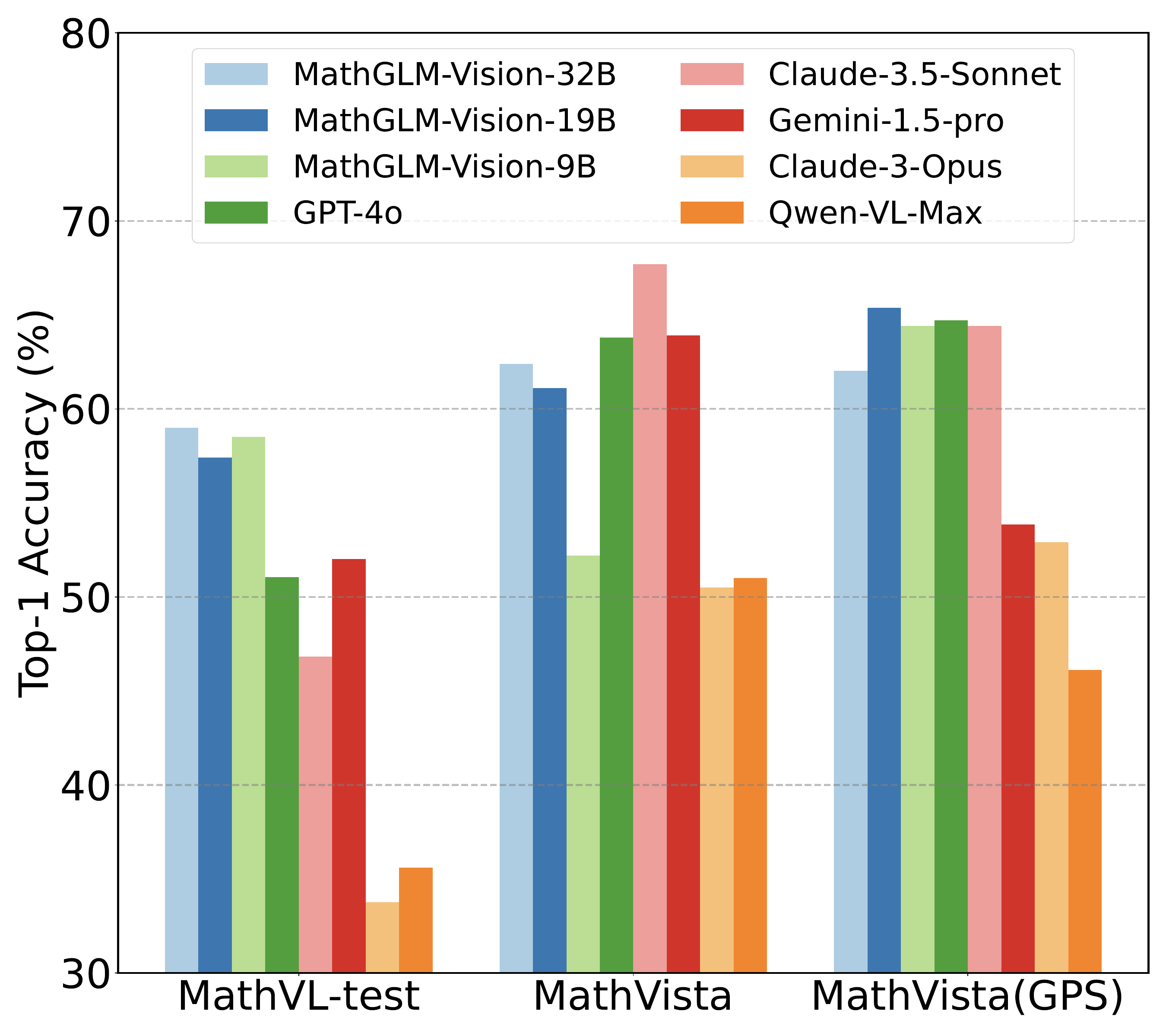}
\end{minipage}%
\begin{minipage}{.57\textwidth}
  \centering
  \includegraphics[width=\linewidth]{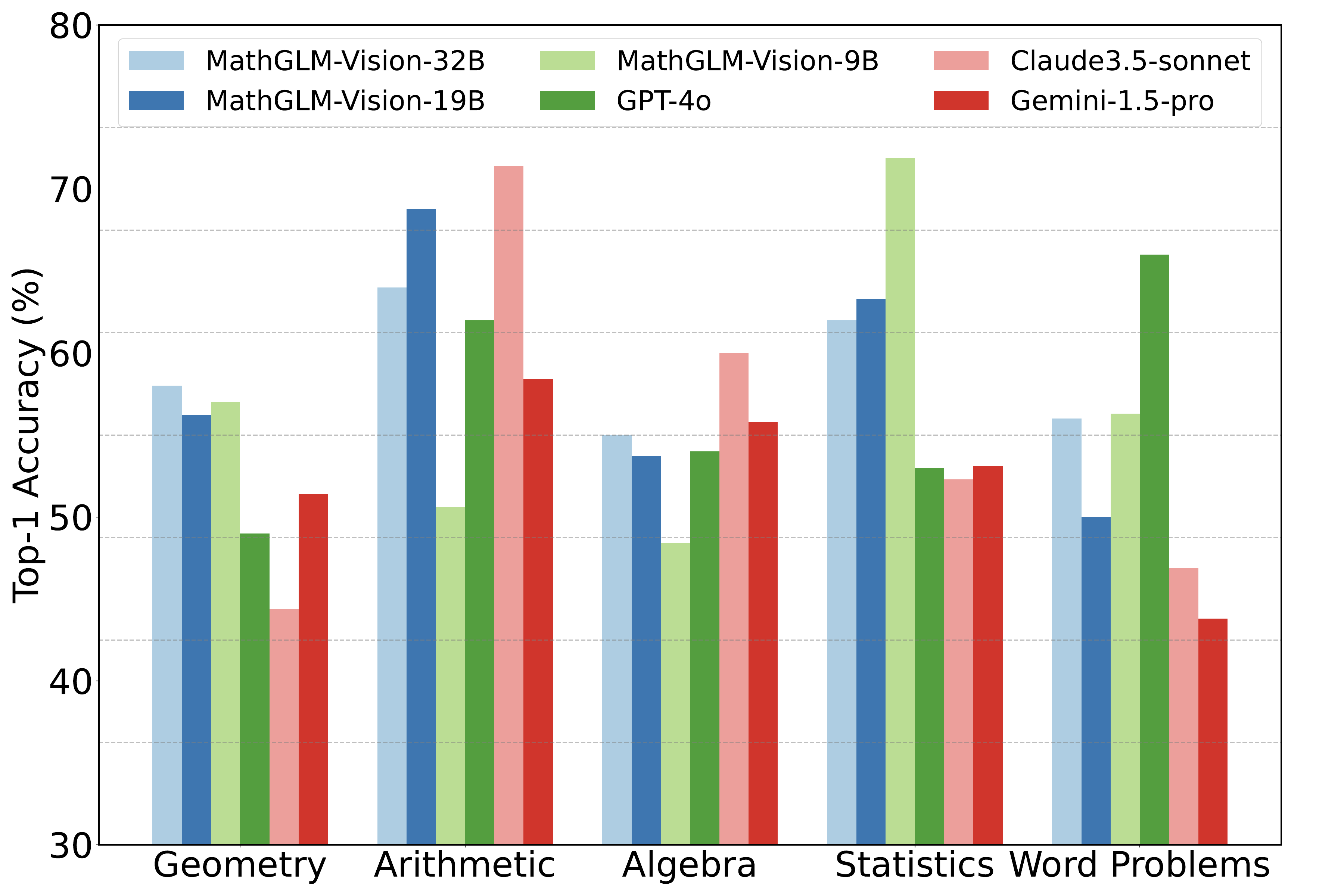}
\end{minipage}
\caption{Performance comparison of the different multi-modal large language models. (Left) The accuracies of \model and other MLLMs among three evaluation datasets. (Right) The accuracy of \model and other MLLMs on \data-test across different categories.}
\label{fig:plot_results}
\end{figure}


In this paper, we construct a fine-tuning dataset named \textbf{\data}, which encompasses both open-source data and our specially curated Chinese data collected from K12 education. The \data dataset is meticulously designed to incorporate a diverse range of mathematical problems, consisting of textual and visual inputs. For textual information, the \data dataset covers a variety of mathematical subjects such as arithmetic, algebra, geometry, statistics, and word problems. It includes various types of questions, including fill-in-the-blank, multiple-choice, and free-form. For visual information, the \data dataset involves elements like functions, statistical data, graphs, charts, LaTeX expressions, and geometric figures, providing a comprehensive resource for complex mathematical problem solving.

With our constructed \data dataset, we develop a series of specialized mathematical MLLMs, collectively referred to as \model, with different parameter scales. Specifically, \model-9B, \model-19B and \model-32B are fine-tuned on three backbone models: GLM-4V-9B, CogVLM2, and CogVLM-32B, respectively. Moreover, we establish a benchmark dataset named \data-test, which contains 2,000 problems designed to evaluate the ability of \model and other MLLMs in solving mathematical problems involving visual information. Through extensive evaluation experiments on four public benchmark datasets and one curated \data-test, we validate the effectiveness of our \model. The results in Figure~\ref{fig:plot_results} demonstrate that \model exhibits superior performance in understanding and solving complex mathematical problems with visual elements compared to existing MLLMs. For instance, on the geometry problem solving (GPS) minitest split of MathVista~\cite{lu2023mathvista}, \model-9B achieves a 39.68\% relative improvement for GLM-4V-9B, \model-19B achieves a 65.06\% relative improvement for CogVLM2, and \model-32B achieves a 51.05\% relative improvement over CogVLM-32B.

\section{MathVL Dataset}

\begin{figure}[hbpt]
    \centering
    \includegraphics[width=1.0\textwidth]{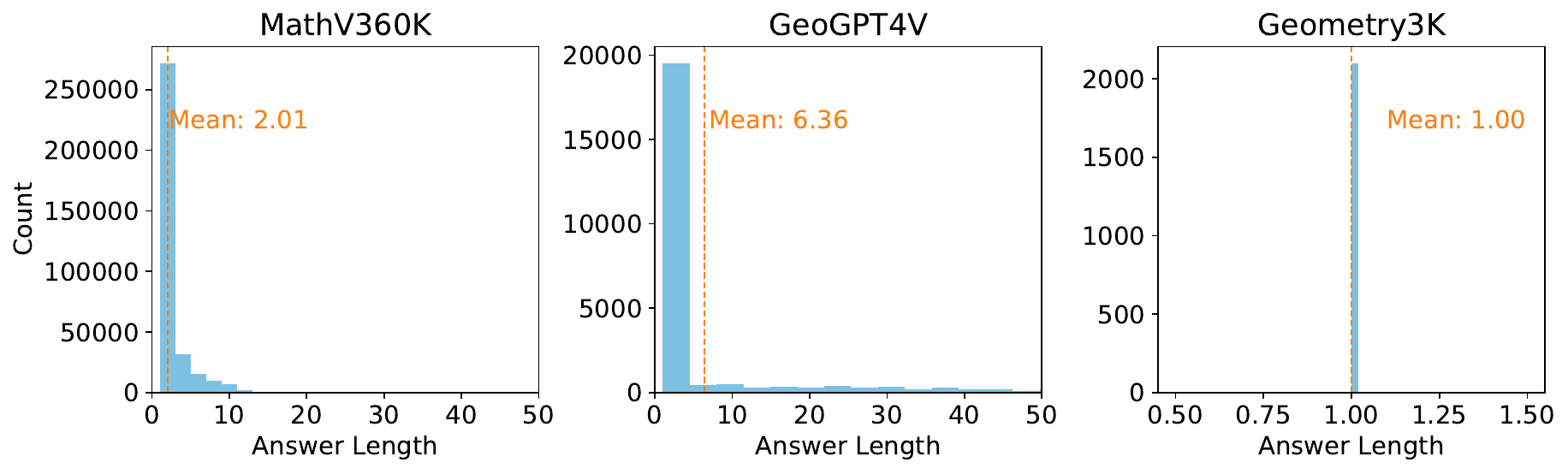}
    \vspace{-2mm}
    \caption{Analysis of answer lengths in several open-source mathematical datasets like MathV360K, GeoGPT4V, and Geometry3K.}
    \label{fig:data_issue}
\end{figure}

To enhance the capabilities of MLLMs in solving mathematical problems, previous efforts~\cite{chen2021geoqa,chen2022unigeo,cao2022augmented,gao2023g} focus on constructing high-quality datasets. Nevertheless, the majority of these datasets fall into the category of Visual Question Answering (VQA), which generally involves descriptive or identification tasks rather than conventional mathematical problems. Furthermore,  the answers in some public datasets like Geometry3K~\cite{lu2021inter}, GeoGPT4V~\cite{cai2024geogpt4v}, MathV360K~\cite{shihu2024mathllava} for standard mathematical questions are often too simplistic, usually providing only the final answer without the intermediate steps necessary for a thorough understanding. It is well-established that including step-by-step solutions can significantly enhance the reasoning capabilities of large language models~\cite{wei2022chain,lightman2023let,zhang2023multimodal,wang2023math}. Figure~\ref{fig:data_issue} demonstrates the distribution of answer lengths in current open-source mathematical datasets. 

To address these issues, we construct a fine-tuning dataset \data, including both several public datasets and our curated Chinese dataset collected from K12 education levels. This dataset is meticulously crafted to encompass a diverse array of mathematical problems that incorporate visual information. Each problem is presented with detailed step-by-step solutions, aiming to enhance the problem-solving skills of MLLMs by providing them with both the context and the procedural knowledge necessary for effective reasoning and comprehension.

\vpara{Open-Source Data.} We first collect open-source datasets from GeoQA+~\cite{cao2022augmented}, Geometry3K~\cite{lu2021inter}, ChartQA~\cite{masry2022chartqa}, and UniGEO-Calculation~\cite{chen2022unigeo}. These datasets commonly serve as seed data for constructing enhanced datasets. Through observation and statistical analysis, we discover that 57\% of the answers within these datasets are comprised of fewer than 50 words, indicating that many questions are answered directly without elaboration or explanation. To enrich these dataset with comprehensive step-by-step solutions, we employ GPT-4o to generate the detailed solutions for each question, thereby enhancing the learning and reasoning potential of these datasets. After generating the detailed answers, we perform a rigorous judgement process to ensure the accuracy of the solutions provided by GPT-4o. Additionally, we adopt a public instruction tuning dataset named Geo170K~\cite{gao2023g}, which is constructed using GeoQA+ and Geometry3K as seed data and contains more than 110K geometric question-answer pairs. We also incorporate another public dataset, GeomVerse~\cite{kazemi2023geomverse}, as part of our resources. In the end, the detailed statistics of the open-source datasets used in \model is provided in Table~\ref{tab:english_dataset}.

\begin{table}[tbhp]
    \centering
    \resizebox{\textwidth}{!}{%
    \begin{tabular}{c|ccccccc}
    \toprule
    Datasets & ChartQA & UniGeo-Calculation & Geometry3K & GeoQA+ & Geo170K & GeomVerse & ALL\\
    \midrule
    Samples & 7,398 & 3,499 & 2,101 & 6,026 & 117,205 & 9,339 & 145,568 \\
    \bottomrule
    \end{tabular}}
    \vspace{0.2cm}
    \caption{The detailed statistics of the open-source datasets used in \model.}
    \label{tab:english_dataset}
    \vspace{-0.5cm}
\end{table}

\begin{figure}[h!]
    \centering
    \includegraphics[width=\textwidth]{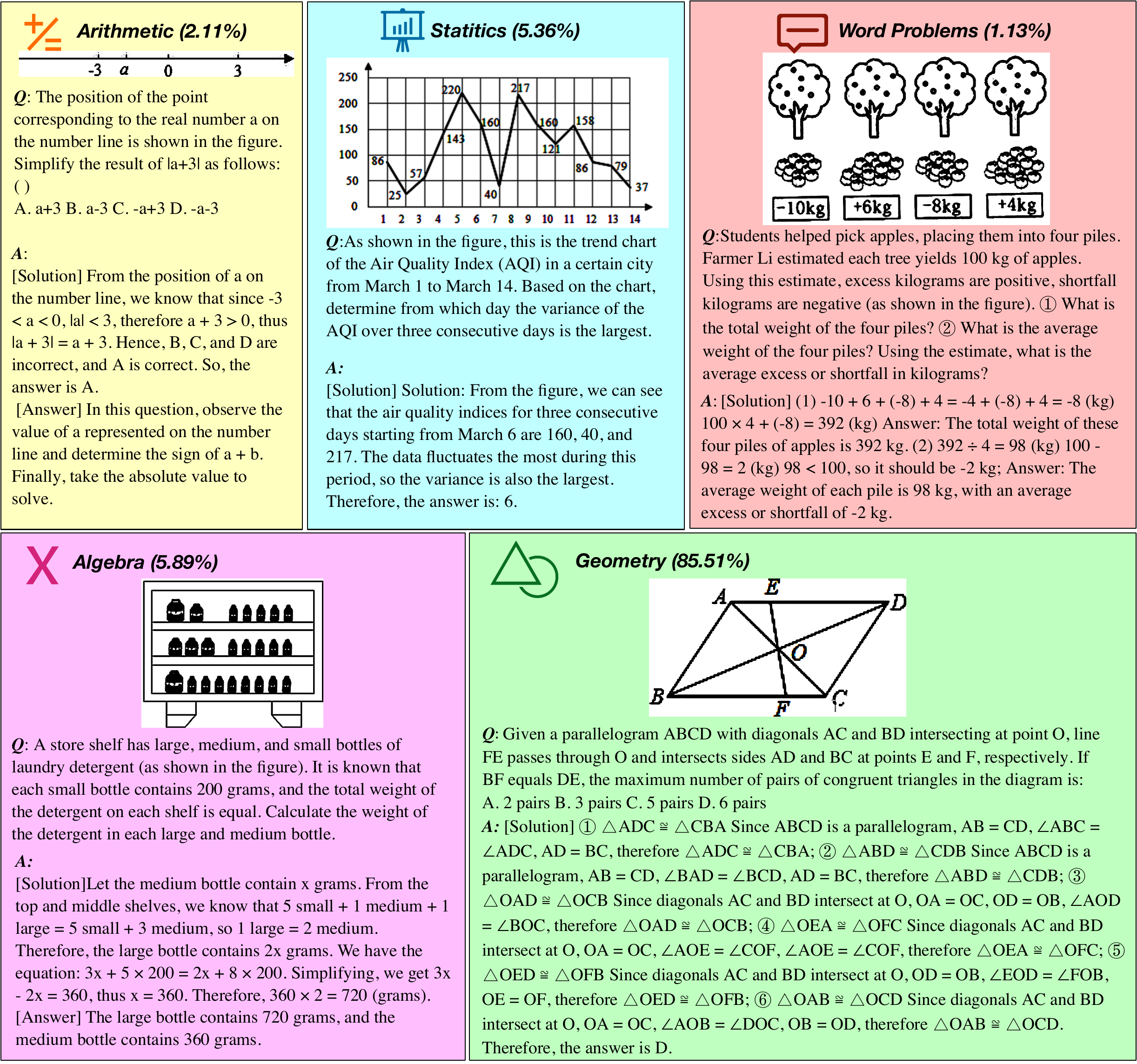}
    \vspace{-2mm}
    \caption{Examples sampled from the constructed Chinese dataset.}
    \label{fig:chinese_examples}
    \vspace{-3mm}
\end{figure}

\vpara{Chinese Data Collected from K12 Education.} We construct a dataset specifically focused on K12 education, comprising 341,346 mathematical problems with textual and visual inputs. This dataset is meticulously curated to encompass a board range of mathematical topics and difficulty levels tailored to the Chinese educational curriculum.  It features various question types such as multiple-choice, fill-in-the-blank, and free-form questions across disciplines including arithmetic, algebra, geometry, statistics, and word problems. Mathematically, this dataset can be represented as $D_{\data}^{zh} = \{Q, A, I_s\}$, where $Q$ represents the question, $A$ represents the answer, and $I_s$ represents one or more images for each corresponding question. To build this dataset, we first process the images by adding a white border around each image and enhancing their resolution to ensure that MLLMs can effectively recognize and interpret these images. This modification is crucial for facilitating the accurate extraction of visual information. Next, we extract 341,346 samples from a raw dataset containing 685,670 samples by implementing a selective filtering process. This selection is based on two specific criteria: (1) filtering out samples where the answer includes images or the question is incomplete, and (2) eliminating samples with answer that are fewer than 50 words in length to ensure the responses are sufficiently detailed for model training. After constructing this dataset, we categorize and analyze it based on mathematical topics associated with each question. Detailed statistics about the distribution of these categories are presented in Table~\ref{tab:chinese_dataset}. Figure~\ref{fig:chinese_examples} demonstrates some examples sampled from the constructed Chinese dataset, providing a visual representation of the mathematical topics of questions included. More dataset cases are provided in Appendix~\ref{appendix: dataset_case}.

\begin{table}[tbhp]
    \centering
    \resizebox{\textwidth}{!}{%
    \begin{tabular}{c|cccccc}
    \toprule
    Types & Arithmetic & Geometry & Algebra & Statistics & Word Problems & ALL\\
    \midrule
    Samples & 7,207 & 291,879 & 20,111 & 18,284 & 3,865 &  341,346 \\
    \bottomrule
    \end{tabular}}
    \vspace{0.2cm}
    \caption{Detailed statistics regarding the distribution  used in \model.}
    \label{tab:chinese_dataset}
    \vspace{-0.5cm}
\end{table}

\section{\model}
\vpara{Model Architecture.} We employ CogVLM~\cite{wang2023cogvlm} and GLM-4V-9B~\cite{glm2024chatglm} architectures as our backbone models, and conduct Supervised Fine-Tuning (SFT) on our constructed \data dataset. Specifically, we utilize three pre-trained multi-modal large language models (MLLMs) for the fine-tuning process: GLM-4V-9B, CogVLM2-19B, and CogVLM-32B. This results in the development of three distinct variants of \model, designated as \model-9B, \model-19B, and \model-32B, respectively. Table~\ref{tab:mathvlm_strcuture} demonstrates an overview of the \model series, detailing the different model parameters and configurations. Further details about the abovementioned three pre-trained MLLMs are available in Appendix~\ref{appendix: backbone_models}.

\begin{table}[tbhp]
    \centering
    \resizebox{\textwidth}{!}{%
    \begin{tabular}{c|c|c|ccc|ccc}
    \toprule
     \multirow{2}{*}{Model} & \multirow{2}{*}{LLM Size} & \multirow{2}{*}{ToTal Size} & \multicolumn{3}{c|}{Language Model} & \multicolumn{3}{c}{Image Encoder}  \\
    & & & Layers & Hidden Size & Heads & Layers & Hidden Size & Heads  \\
    \midrule
    \model-9B & 9B & 20B &40  & 4096  & 32 & 63 & 1792 & 16 \\
    \model-19B & 8B & 19B & 32 & 4096 & 32 & 63 & 1792 & 16  \\
    \model-32B & 32B & 43B & 58  & 6656 & 52 & 63 & 1792 & 16   \\
    \bottomrule
    \end{tabular}}
    \vspace{0.2cm}
    \caption{An overview of \model series along with model parameters and configurations.}
    \label{tab:mathvlm_strcuture}
    \vspace{-0.5cm}
\end{table}

\vpara{Model Training.} To maintain the general vision-language understanding skills of \model, we incorporate 19 open-source visual question-answering datasets (VQA datasets) into the \data dataset. More details about the task type and visual context of VQA datasets are provided in Appendix~\ref{appendix: vqa_datasets_details}. These datasets are meticulously selected to challenge and enhance the model's ability to interpret and integrate visual and textual information, ensuring it retains a broad understanding across various contexts. By merging these varied sources, we enhance \model's specialized capabilities for mathematical problem-solving and simultaneously preserve its robustness in general vision-language tasks. In the end, we conduct supervised fine-tuning (SFT) across the combined VQA and \data datasets. The training process undergoes 35,000 iterations with a learning rate of 1e-5 and a batch size of 128. To ensure the stability of the training, we activate the visual encoder's parameters and adjust its learning rate to be one-tenth of that used for the remaining training parameters. The details of the SFT procedures are described in Appendix~\ref{appendix: finetuning_setup}.


\section{Experiments}
\subsection{Experimental Setup}

\vpara{Evaluation Datasets.} We assess our \model using three well-established public benchmark datasets(MathVista~\cite{lu2023mathvista}, MathVerse~\cite{zhang2024mathverse}, and Math-Vision~\cite{wang2024measuring} datasets) alongside our specially curated dataset \data-test. The \data-test dataset comprises 2,000 sampled cases, distinct from those in the \data dataset, ensuring a rigorous and unbiased evaluation of \model's capabilities. Besides, we adopt the testing protocol~\cite{gao2023g} and utilize MathVista-GPS dataset to evaluate \model's ability to solve geometry problems. Additionally, we evaluate \model's general vision-language understanding skills using the MMMU benchmark~\cite{yue2024mmmu}. Detailed descriptions for these benchmark datasets are provided in Appendix~\ref{appendix: benchmark_dataset}.

\vpara{Compared Models.} We compare \model with other Multi-Modal Large Language Models (MLLMs), including closed-source MLLMs such as Gemini~\cite{team2023gemini}, GPT-4V~\cite{openai2023gpt4v}, Claude3~\cite{Claude3}, and Qwen-VL~\cite{bai2023qwen-vl}, and open-source MLLMs like mPLUG-Owl~\cite{ye2023mplug}, LLaMA-Adapter-V2~\cite{gao2023llama}, InstrctBLIP~\cite{dai2024instructblip}, LLaVA-1.5~\cite{liu2024llava}, ShareGPT4V~\cite{chen2023sharegpt4v}, SPHINX~\cite{gao2024sphinx}, InternLM-XC2~\cite{dong2024internlm}, and InternVL~\cite{chen2024internvl}. Additionally, we compare \model with recent specialized mathematical MLLMs, including G-LLaVA~\cite{gao2023g}, LLaVA-1.5-G~\cite{cai2024geogpt4v}, ShareGPT4V-G~\cite{cai2024geogpt4v}, and Math-LLaVA~\cite{shihu2024mathllava}.

\vpara{Evaluation Metrics.} We adopt top-1 accuracy to evaluate our \model across MathVista-GPS, MathVista, MathVerse, MathVision, and \data-test. Our evaluation process follows the pipeline outlined in the aforementioned benchmark datasets, which involves using LLMs to extract predicted answers from the model’s responses. Accuracy is then calculated by comparing these extracted answers against the ground truths.

\subsection{Main Results}

\vpara{Results on public benchmark datasets.} To comprehensively assess the ability of \model in solving mathematical problems, we evaluate its performance against other MLLMs across several public benchmark datasets, including MathVista-GPS~\cite{lu2023mathvista}, the testmini subset of MathVista~\cite{lu2023mathvista}, MathVerse~\cite{zhang2024mathverse}, and Math-Vision~\cite{wang2024measuring}. Table~\ref{tab:results_on_benchmarks} demonstrates the overall results from these evaluations. The experimental results indicate that our constructed \data dataset can significantly improve \model's mathematical reasoning capabilities. For example, \model-9B achieves a 64.42\% accuracy on the MathVista-GPS dataset, marking a substantial 39.68\% improvement over its backbone model, GLM-4V-9B. Besides, across various parameter scales, \model consistently surpass all backbone models on different evaluation benchmarks, highlighting the significant enhancements that \data brings to the \model's problem-solving skills. Notably, \model outperforms all open-source specialized mathematical MLLMs across various benchmarks. The superior performance suggests that the high-quality and diverse data, complete with detailed step-by-step solutions, are crucial for improving MLLM's mathematical reasoning capabilities. More importantly, \model-32B outperforms even the advanced GPT-4V on the more challenging Math-Vision benchmark, demonstrating its superior capacity to tackle complex mathematical problems. Detailed experimental results on public benchmark datasets across different task types can be found in Appendix~\ref{appendix:detailed_public}.

\begin{table*}[htbp]
    \centering
    \renewcommand{\arraystretch}{1.15}
    \small
    \resizebox{\textwidth}{!}{%
    \begin{tabular}{ccccccc}  
    \toprule
     Model & Input & LLM   & MathVista (GPS) & MathVista & MathVerse  & MathVision  \\
    \midrule
    \multicolumn{7}{c}{\textit{Closed Source Models}}\\
     \midrule
     Gemini Pro &    \textit{Q}, \textit{I} & - & 40.40  & 45.20 &  36.80  & 17.66 \\ 
     Gemini-1.5-Pro &    \textit{Q}, \textit{I} & - & 53.85 & 63.90  & 51.08 &  19.24  \\ 
     GPT-4V & \textit{Q}, \textit{I} & - & 50.50  & 49.90  &  50.80 & 22.76     \\ 
     GPT-4-turbo & \textit{Q}, \textit{I} & - &  58.25   & 58.10 & 43.50 & 30.26  \\ 
     GPT-4o & \textit{Q}, \textit{I} & - & \cellcolor{red!25}{64.71} & 63.80  & \cellcolor{red!25}{56.65}  & 30.39  \\ 
     Claude3-Opus & \textit{Q}, \textit{I}  & - & 52.91 & 50.50 & 31.77 & 27.13  \\
     Claude3.5-Sonnet & \textit{Q}, \textit{I}  & - & 64.42 & \cellcolor{red!25}{67.70} & 48.98 &  \cellcolor{red!25}{37.99} \\
     Qwen-VL-Plus & \textit{Q}, \textit{I} & - & 33.01 & 43.30 & 19.10 & 10.72    \\
     Qwen-VL-Max & \textit{Q}, \textit{I} & - & 46.12 & 51.00 & 35.90 &  15.59   \\
     \midrule
    \multicolumn{7}{c}{\textit{Open Source Models }}\\
    \midrule
    \multicolumn{7}{l}{\textit{General Multi-modal LLMs}}\\
    mPLUG-Owl  & \textit{Q}, \textit{I} & LLaMA-7B & 23.60 & 22.20 & 12.47 &   9.84 \\
    LLaMA-Adapter-V2  & \textit{Q}, \textit{I} & LLaMA-7B  & 25.50 & 23.90 & 4.50  & 9.44   \\
    InstructBLIP  & \textit{Q}, \textit{I} & Vicuna-7B & 20.70 & 25.30 & 15.36 &  10.12 \\
    LLaVA-1.5  & \textit{Q}, \textit{I} & Vicuna-13B & 24.04  &  27.60 & 12.70 & 11.12   \\
    ShareGPT4V & \textit{Q}, \textit{I} & Vicuna-13B &  38.35 & 29.30 & 16.20 & 11.88   \\
    SPHINX-MoE & \textit{Q}, \textit{I} & Mixtral 8*7B & 31.20 & 42.30  & 19.60 & 14.18\\
    SPHINX-Plus & \textit{Q}, \textit{I} & LLaMA2-13B & 16.40  & 36.70 & 14.70 & 9.70 \\
    InternLM-XC2  & \textit{Q}, \textit{I} & InternLM2-7B &63.00 & 57.60 & 24.40 & 14.54  \\
    InternVL-1.2-Plus & \textit{Q}, \textit{I} & 
Nous-Hermes-2-Yi-34B & 61.10 & 59.90 & 21.70 & 16.97   \\
    \midrule
    \multicolumn{7}{l}{\textit{Geo-Multi-modal LLMs}}\\
    G-LLaVA & \textit{Q}, \textit{I} & LLaMA2-7B & 53.40 & 28.46 & 12.70 & 12.07   \\
    G-LLaVA & \textit{Q}, \textit{I} & LLaMA2-13B & 56.70 & 35.84 & 14.59  & 13.27    \\
    LLaVA-1.5-G & \textit{Q}, \textit{I} & Vicuna-7B & 32.69 & 45.22   & 13.96  & 14.13   \\
    LLaVA-1.5-G & \textit{Q}, \textit{I} & Vicuna-13B  & 36.54 & 48.34 & 15.61 & 14.88 \\
    ShareGPT4V-G & \textit{Q}, \textit{I} & Vicuna-7B & 32.69 & 45.07 & 16.24 & 12.86 \\
    ShareGPT4V-G & \textit{Q}, \textit{I} & Vicuna-13B & 43.27 & 49.14 & 16.37 & 14.45 \\
    Math-LLaVA  & \textit{Q}, \textit{I} & Vicuna-13B &  57.70 & 46.60 & 19.04 & 15.69  \\
    \midrule
    \multicolumn{7}{c}{\textit{\model and Backbone Models}}\\
    \midrule
    GLM-4V-9B  & \textit{Q}, \textit{I} & GLM-4-9B  & 46.12 & 46.70   & 35.66  &  15.31 \\
    \textbf{\model-9B} & \textit{Q}, \textit{I}  & GLM-4-9B & \textbf{64.42}  & \textbf{52.20} &  \textbf{44.20}  & \textbf{19.18}   \\
    \hline
    CogVLM2  & \textit{Q}, \textit{I} & LLaMA-3-8B & 39.61  & 40.85 &  25.76 &  13.20 \\
    \textbf{\model-19B} & \textit{Q}, \textit{I} & LLaMA-3-8B & \textbf{65.38} & \textbf{61.10} & \textbf{42.50} & \textbf{21.64} \\
    \hline
    CogVLM-32B  & \textit{Q}, \textit{I}  & GLM2-32B & 41.06 & 40.04   & 35.28  &  19.32 \\
    \textbf{\model-32B} & \textit{Q}, \textit{I} & GLM2-32B & \textbf{62.02}  & \textbf{62.40} & \textbf{49.20} & \textbf{26.51}  \\
    \bottomrule
    \end{tabular}} 
    \caption{\textbf{Results on several public benchmark datasets.} Comparison of model performance on the testmini set of MathVista and geometry problem solving (GPS) of MathVista. For MathVerse dataset, results are evaluated on Vision Dominant with CoT-E. For Math-Vision dataset, all 3,040 samples included in the data are evaluated. The highest accuracy among all baseline MLLMs is marked in red, while the highest accuracy among various variants of \model is marked bold.}
    \label{tab:results_on_benchmarks}
    \vspace{-0.48cm}
\end{table*}

\vpara{Results on \data-test.} We also evaluate \model and several close-source MLLMs using our specially constructed \data-test dataset. Given that \data-test comprises multiple images for each question, we exclusively assess the performance of these close-source MLLMs through API calls. As depicted in Table~\ref{tab:main_results_mathvl}, the results clearly demonstrate that \model significantly outperforms both its backbone models and other leading closed-source MLLMs across various model sizes. Specifically, our \model-32B outperforms the advanced GPT-4o with a significant margin, achieving an accuracy of 59.00\% compared to GPT-4o's 51.05\%. Compared to the backbone model, GLM-4V-9B, \model-9B achieves an impressive accuracy of 57.05\% with a significant improvement of 86.5\%. This superior performance suggests that \model, when conducting SFT on the \data dataset, notably enhances its capability to tackle complex Chinese mathematical problems. Additionally, we report the accuracy across various categories, as illustrated in Figure~\ref{fig:plot_results} (See Right). \model significantly outperforms other advanced MLLMs in the domains of geometry and statistics. In contrast, Claude3.5-Sonnet excels in algebra and arithmetic, demonstrating superior performance. Meanwhile, \model-19B ranks second in performance in the domain of arithmetic, showing its strong abilities in this area as well. GPT-4o exhibits the highest performance in word problems domain, while \model also exhibits robust performance, surpassing both Gemini-1.5-Pro and Claude3.5-Sonnet in this category.

\begin{table*}[htbp]
    \centering
    \renewcommand{\arraystretch}{1.15}
    \small
    \resizebox{0.6\textwidth}{!}{%
    \begin{tabular}{cccc}  
    \toprule
     Model & Input & LLM Size  & \data-test \\
     \midrule
     Gemini-1.5-Pro & \textit{Q}, \textit{I} & - & \cellcolor{red!25}{52.03} \\
     GPT-4V & \textit{Q}, \textit{I} & -    & 35.89   \\ 
     GPT-4-turbo & \textit{Q}, \textit{I} & -  & 42.19  \\ 
     GPT-4o & \textit{Q}, \textit{I} & -  &  51.05 \\
     Claude3.5-Sonet & \textit{Q}, \textit{I} & -  &  46.84 \\
     Claude3-Opus & \textit{Q}, \textit{I} & -  &  33.77 \\
     Qwen-VL-Plus & \textit{Q}, \textit{I} & -  & 28.50\\
     Qwen-VL-Max & \textit{Q}, \textit{I} & -  & 35.61 \\
     \midrule
    GLM-4V-9B  & \textit{Q}, \textit{I} & 9B   & 30.59 \\
    \textbf{\model-9B} & \textit{Q}, \textit{I}  & 9B & \textbf{57.05} \\
    \hline
    CogVLM2  & \textit{Q}, \textit{I} & 8B & 27.47 \\
    \textbf{\model-19B} & \textit{Q}, \textit{I} & 8B  & \textbf{57.30} \\
    \hline
    CogVLM-32B  & \textit{Q}, \textit{I}  & 32B  & 30.86 \\
     \textbf{\model-32B} & \textit{Q}, \textit{I} & 32B & \textbf{59.00}  \\
    \bottomrule
    \end{tabular}} 
    \caption{\textbf{Results on \data-test.} A detailed comparison of the performance of \model and various other leading close-source MLLMs on the \data-test. The highest accuracy among all baseline MLLMs is marked in red, while the highest accuracy among various variants of \model is marked bold.}
    \label{tab:main_results_mathvl}
\end{table*}

\subsection{Generalizability of \model}

In addition to its proficiency in mathematical reasoning, we further assess \model's capabilities in general vision-language understanding by conducting experiments on the MMMU benchmark~\cite{yue2024mmmu}. This benchmark is specifically designed to evaluate the ability of models to comprehend and process information across a variety of academic and professional disciplines, providing a comprehensive test of general vision-language understanding. Table~\ref{tab:general_results} shows the performance of \model, a specific variant fine-tuned exclusively on \data without the inclusion of VQA datasets, and backbone models. Compared to CogVLM2, \model-19B achieves comparable performance in terms of generalizability, underscoring its capacity for simultaneous multi-modal understanding and mathematical reasoning. However, \model-32B shows a slight reduction in performance across multiple categories on the MMMU benchmark. Besides, \model, when fine-tuned with VQA datasets, outperforms its variant lacking VQA datasets. This indicates that omitting VQA datasets from the fine-tuning process limits the general vision-language understanding abilities. Thus, the Specialized Fine-Tuning (SFT) process using our \data incorporated with VQA datasets not only enhances \model's mathematical reasoning abilities but also preserves its generalizability.

\begin{table}[hbpt]
    \centering
    \renewcommand{\arraystretch}{1.15}
    \resizebox{\textwidth}{!}{%
    \begin{tabular}{c|c|cccccc}  
    \toprule
    Model  & MMMU & Art \& Design & Business & Sci. & Health \& Med. & Human. \& Social Sci. & Tech. \& Eng.  \\
    \midrule
    CogVLM2 & 40.2 & 58.3 & 30.0 & 26.7 & \textbf{41.3} & 38.6 & \textbf{53.3} \\
    w/o VQA datasets & 38.1 & 60.8 & 28.7 & \textbf{34.0} & 36.7 & 43.3 & 32.9 \\
    \model-19B & \textbf{40.2} & \textbf{63.3} & \textbf{37.3} & 27.3 & 36.0 & \textbf{46.7} & 37.6 \\
    \midrule
    CogVLM-32B & \textbf{42.9} & \textbf{63.3} & \textbf{31.3} & 32.0 & \textbf{43.3} & \textbf{62.5} & \textbf{36.7}  \\
    w/o VQA datasets & 38.6 & 62.5 & 26.7 & 28.0 & 34.0 & 56.7 & 33.8 \\
    \model-32B & 40.0 & 60.0  & 28.7 & \textbf{34.0} & 38.7  & 52.5  & 34.8 \\
    \bottomrule
    \end{tabular}} 
    \vspace{0.2cm}
    \caption{Generalizability of \model on the MMMU benchmark.}
    \label{tab:general_results}
    \vspace{-0.3cm}
\end{table}

\subsection{Further Analysis}

\vpara{Effect of Chinese Dataset.} To validate the effectiveness of the adopted Chinese dataset in \data, we conduct an extend experiment that involves fine-tuning GLM-4V-9B with open-source datasets, deliberately excluding Chinese data collected from K12 education. The purpose of this experiment is to assess the specific contributions of the Chinese dataset to the capabilities of \model. Table~\ref{tab:effect_data} shows a comparison of performance results. Compared to the backbone model GLM-4V-9B, a variant \model-9B that undergoes Specialized Fine-Tuning (SFT) exclusively with open-source data exhibits significant improvement on the minitest of MathVista, particularly in geometry problem solving (GPS) and geometry reasoning (GEO). This indicates that fine-tuning on diverse open-source data can markedly enhance model performance in specific mathematical areas. \model, incorporating both open-source data and Chinese data, outperforms the variant tuned only with open-source data on the minitest of MathVista, highlighting the significant value added by integrating the Chinese dataset in the training process. Notably, compared to the variant without Chinese data, \model achieves a significantly higher accuracy on the \data-test. These findings confirm that the inclusion of the Chinese dataset not only enhances the model's capability in handling complex mathematical problems but also contributes significantly to its overall performance on a diverse set of tasks within MathVista.

\vpara{Effect of VQA Datasets.} To explore the effect of VQA datasets on the performance of \model, an extended experiment can be designed where Specialized Fine-Tuning (SFT) is applied exclusively to mathematical datasets, deliberately excluding VQA datasets. Table~\ref{tab:vqa_dataset} demonstrates the performance comparison achieved by different models on MathVista. Compared to the backbone model GLM-4V-9B, a variant of \model-9B achieves significant improvements on geometry problem solving (GPS) and geometry Reasoning (GEO). However, it exhibits a decline in the overall accuracy on the minitest of MathVista (ALL). The decline can be attributed to the composition of MathVista, which comprises five tasks, with question-answering types (such as graphical question-answering, textbook question-answering, and visual question-answering) comprising up to 60.6\% of the tasks. Omitting VQA training in \model impacts the model's ability to effectively process and respond to these multi-modal questions. Notably, within specific subsets of MathVista, such as GPS and GEO, a variant of \model-9B slightly below the standard \model-9B. This observation suggests that VQA datasets are crucial for preserving overall multi-modal understanding, their impact may vary depending on different task types. Besides, VQA datasets can indirectly bolster mathematical reasoning skills, which in turn enhances image recognition capabilities.

\begin{table}\footnotesize
\begin{minipage}[t]{0.5\textwidth}
\centering
\small
\setlength{\tabcolsep}{.5mm}{
\renewcommand\arraystretch{1.2}
\begin{small}
\begin{tabular}{c|ccc|c}
    \toprule
     \multirow{2}{*}{Model}   & \multicolumn{3}{c|}{MathVista} & \multirow{2}{*}{\data-test} \\
     & GPS & GEO   & ALL & \\
      \midrule
      GLM-4V-9B  & 46.12 & 44.35  & 46.70  & 30.59 \\
      + SFT on Open-source Data & 62.98 & 61.51 & 50.40  & 47.55 \\
      \midrule
      \model-9B   & 64.42 & 62.34  & 52.20 & 57.25 \\
    \bottomrule
\end{tabular} 
\vspace{0.2cm}
\caption{Effect of the constructed Chinese data.}
\label{tab:effect_data}
\end{small}}
\end{minipage}\hfill
\begin{minipage}[t]{0.38\textwidth}
\centering
\small
\setlength{\tabcolsep}{.5mm}{
\renewcommand\arraystretch{1.2}
\begin{small}
   \begin{tabular}{c|ccc}
    \toprule
    \multirow{2}{*}{Model}  & \multicolumn{3}{c}{MathVista} \\
       & GPS & GEO & ALL \\
      \midrule
      GLM-4V-9B & 46.12 & 44.35 & 46.70 \\
      \midrule
      \model-9B   & 64.42 & 62.34 & 52.20  \\
       - SFT on VQA Datasets & 61.54 & 58.58 & 41.34 \\
    \bottomrule
    \end{tabular} 
     \vspace{0.2cm}
    \caption{Effect of the VQA datasets.}
    \label{tab:vqa_dataset}
\end{small}}
\end{minipage}
\end{table}



\subsection{Error Analysis}

\begin{figure}[h]
\centering
\begin{minipage}{.48\textwidth}
  \centering
  \includegraphics[width=0.98\linewidth]{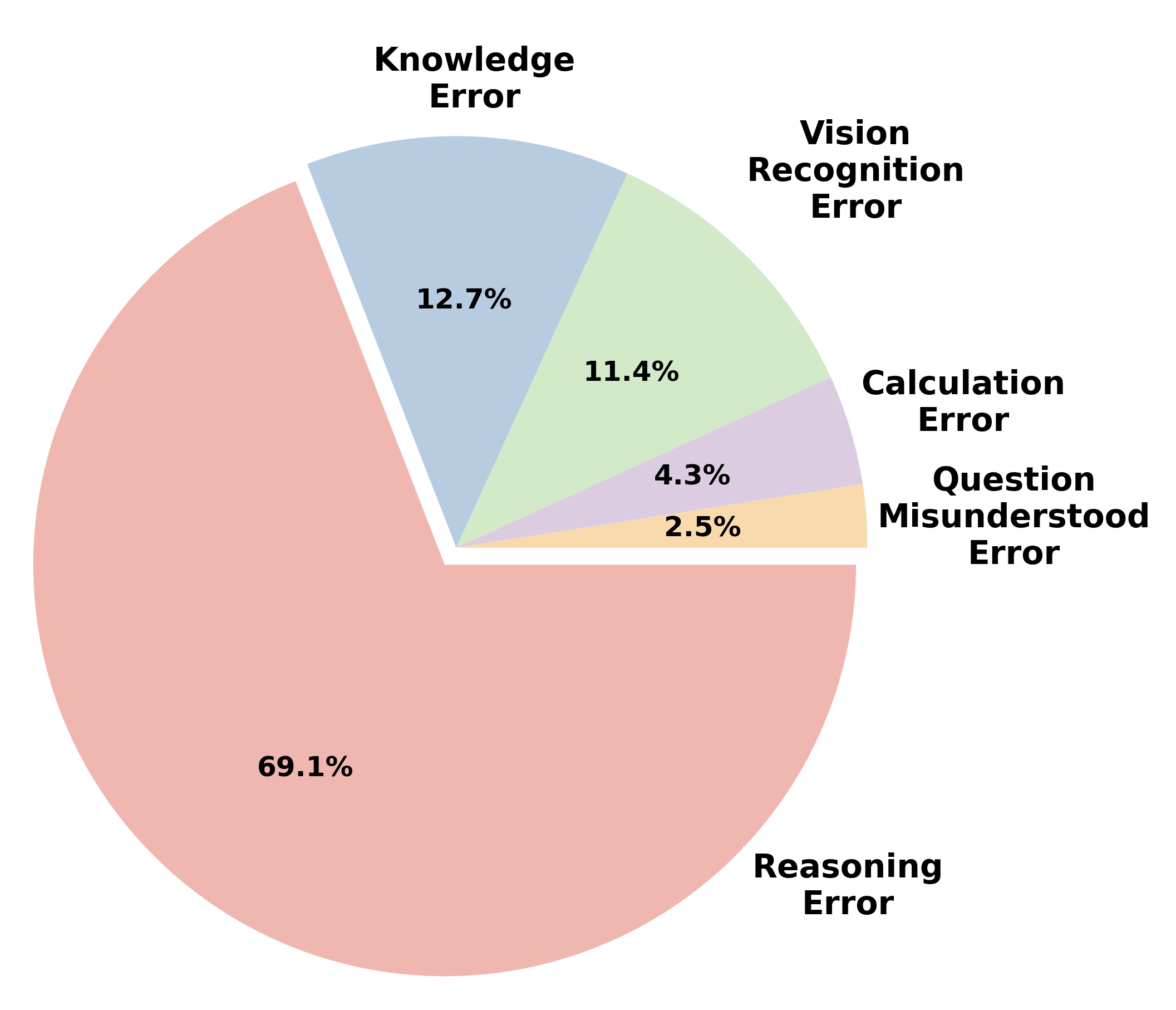}
  \captionof{figure}{Error distribution of MathGLM-\\Vision-32B.}
  \label{fig:error_dist}
\end{minipage}%
\begin{minipage}{.48\textwidth}
  \centering
  \includegraphics[width=0.98\linewidth]{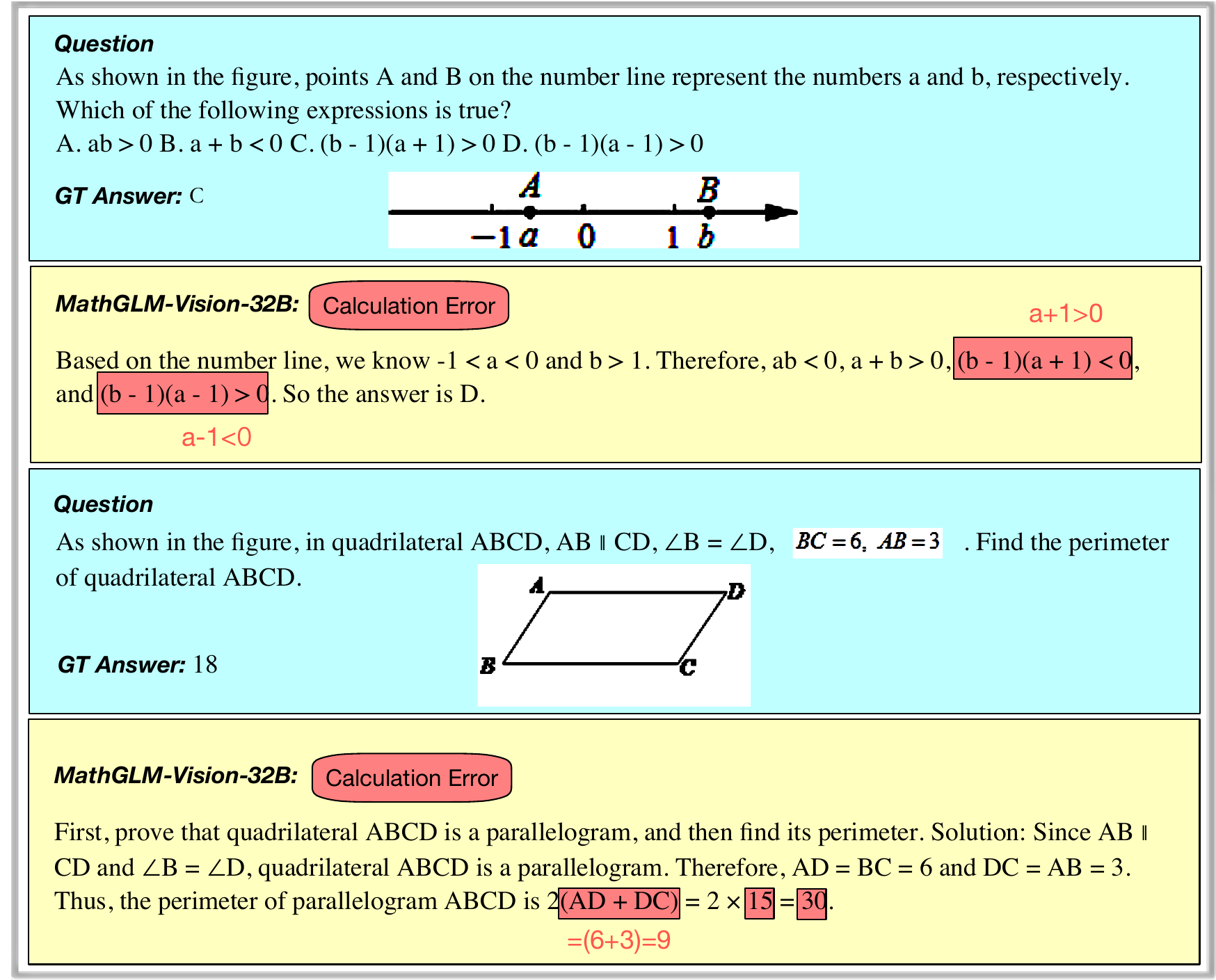}
  \captionof{figure}{Cases of Calculation Error category.}
  \label{fig:error_one_case}
\end{minipage}
\end{figure}

We meticulously analyze the causes of errors in \model-32B on the \data-test dataset and illustrate the distribution of these errors in Figure~\ref{fig:error_dist}. We summarize these errors in \model-32B into five types: reasoning error, knowledge error, vision recognition error, calculation error, and question misunderstood error. The most prevalent type of errors, accounting for 69.1\% of the total, is identified as Reasoning Error. This indicates a significant challenge in the \model-32B's logical deductions and inferential reasoning. Improving these capabilities can dramatically enhance the \model-32B's overall performance. Knowledge Error, which made up 12.7\% of the errors, relates to the model's misapplicaion or lack of specific factual information. Vision Recognition Error accounts for 11.4\% of the total errors and involves inaccuracies in interpreting visual data. This type of error can be reduced through the implementation of more advanced vision encoders. Furthermore, the fact that Calculation Error constitutes only 4.3\% of the errors suggests that \model-32B demonstrates considerable robustness in numerical and computational tasks. Lastly, Question Misunderstood Error, which constitutes 2.5\% of the total, occurs when the model fails to correctly interpret question. Enhancing natural language processing capabilities and refining context understanding can significantly reduce these types of errors. Addressing these identified error types through targeted enhancements can significantly boost the overall effectiveness of \model-32B. Figure~\ref{fig:error_one_case} demonstrate some cases of the Calculation Error category. More detailed examples of these errors can be found in Appendix~\ref{appendix: error_cases}.

\section{Related Works}
\subsection{Multi-Modal Language Model}

The development of Multi-Modal Language Models (MLLMs) have emerged as a significant area of research, which are designed to integrate information from multiple modalities—typically text and images—to perform tasks that require a holistic understanding of both visual and linguistic inputs. Pioneering efforts such as ViLBERT~\cite{lu2019vilbert} and LXMERT~\cite{tan2019lxmert} have advanced this field by conducting the joint pre-training on image-text datasets. They process text and image inputs separately before fusing them for final task layers, significantly improving performance on tasks like image captioning and visual question answering. The continues evolution of MLLMs has lead to innovations in data fusion techniques. Notable models such as CLIP~\cite{radford2021learning}, ALIGN~\cite{jia2021scaling}, and BLIP~\cite{li2022blip} have adopted contrastive learning paradigms to align visual and language information from billions of image-text pairs. Concurrently, the success of LLMs~\cite{brown2020language,du2021glm,zeng2022glm,le2023bloom,bai2022training,touvron2023llama1,ouyang2022training,hoffmann2022training,smith2022using,chowdhery2023palm} facilitates the integration of LLMs into multi-modal tasks by utilizing pre-training alignment and visual instruction tuning, leading to the emergence of multi-modal language models (MLLMs)~\cite{liu2024visual,liu2024llava,wang2023cogvlm,li2023blip,dai2024instructblip,bai2023qwen}. 
Despite MLLMs have demonstrated remarkable capabilities on tasks such as image caption and visual question answering, they stall face significant challenges in solving mathematical problems that involve visual information~\cite{yue2024mmmu,lu2023mathvista,zhang2024mathverse,wang2024measuring}. 

\subsection{Mathematical Reasoning}
Recently, math-specific LLMs~\cite{azerbayev2023llemma,wang2023mathcoder,yue2024mammoth2,ying2024internlm,yu2023metamath,yue2023mammoth,yuan2023scaling,luo2023wizardmath} have demonstrated remarkable abilities in handling mathematical reasoning tasks that only involve textual information. These models have been specifically trained on web-scale instruction mathematical dataset or fine-tuned on specialized mathematical problem sets. For instance, WizardMath~\cite{luo2023wizardmath} and MetaMath~\cite{yu2023metamath} have implemented data augmentation methods to enhance the models' ability to understand and solve mathematical problems by enriching the MATH~\cite{hendrycks2021measuring} and GSM8K~\cite{cobbe2021training} datasets. Recent research has also focused on creating specialized MLLMs for mathematical tasks. UniGeo~\cite{chen2022unigeo} and UniMath~\cite{liang2023unimath} have demonstrated enhanced datasets and conventional deep learning approaches for geometric problem solving. MLLMs like G-LLaVA~\cite{gao2023g}, GeoGPT4V~\cite{cai2024geogpt4v}, and Math-LLaVA~\cite{shihu2024mathllava} are tailored for mathematical problem solving, incorporating both geometric understanding and algebraic reasoning. Additionally, several benchmark datasets~\cite{yue2024mmmu,lu2023mathvista,zhang2024mathverse,wang2024measuring} are proposed to evaluate the multi-modal mathematical reasoning abilities of MLLMs.

\section{Conclusion}
In this paper, we attempt to address the issues in current mathematical MLLMs. We construct a fine-tuning dataset named \data, upon which we conduct a Supervised Fine-Tuning (SFT) process. This initiative results in the development of a series of enhanced MLLMs, designated as \model. Specially, \model contains three variations: \model-9B, \model-19B, and \model-32B, each fine-tuned on different backbone models: GLM-4-V, CogVLM2, and CogVLM-32B, respectively. These developed \model significantly improve the capabilities of mathematical reasoning, achieving substantial performance improvements. Relative to their respective backbone models, \model-9B, \model-19B, and \model-32B show improvements of 39\%, 65\%, and 53.7\% on the Geometry Problem Solving (GPS) minitest split of MathVista, demonstrating the effectiveness of \data in enhancing the mathematical problem-solving abilities of MLLMs. Additionally, we evaluate the effectiveness of \model on our curated \data-test dataset. Experimental results reveal that \model not only surpass their backbone models in specialized mathematical tests but also preserve the generalizability capabilities in general vision-language understanding domains.

{\small
\bibliographystyle{plainnat}
\bibliography{reference}
}

\appendix

\clearpage

\section{Dataset Cases}\label{appendix: dataset_case}

\begin{figure}[hbpt]
    \centering
    \includegraphics[width=1.0\textwidth]{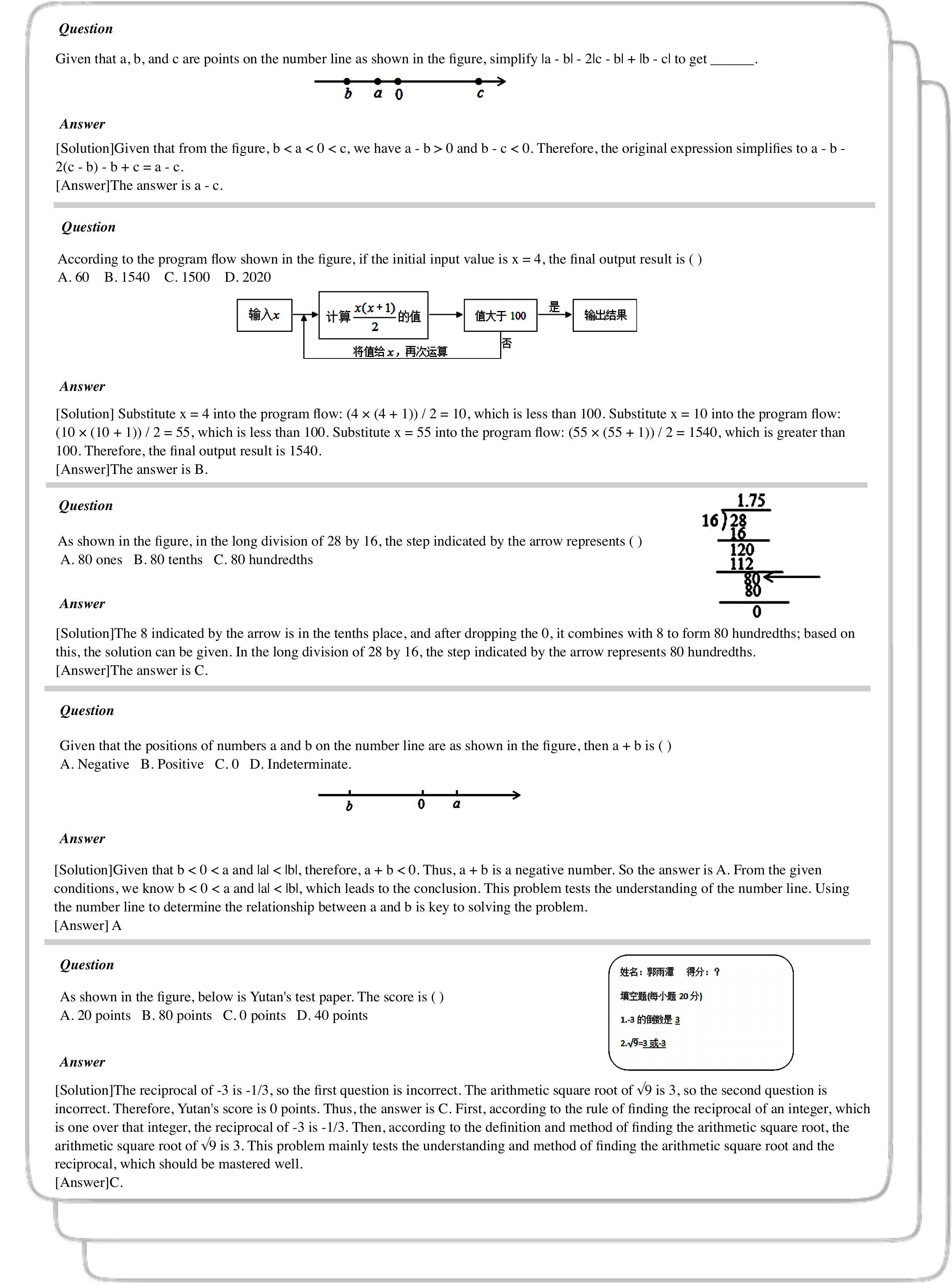}
    \vspace{-2mm}
    \caption{Cases of arithmetical problems in our \data dataset.}
    \label{fig:dataset_arithmetic}
    \vspace{-3mm}
\end{figure}

\begin{figure}[hbpt]
    \centering
    \includegraphics[width=1.0\textwidth]{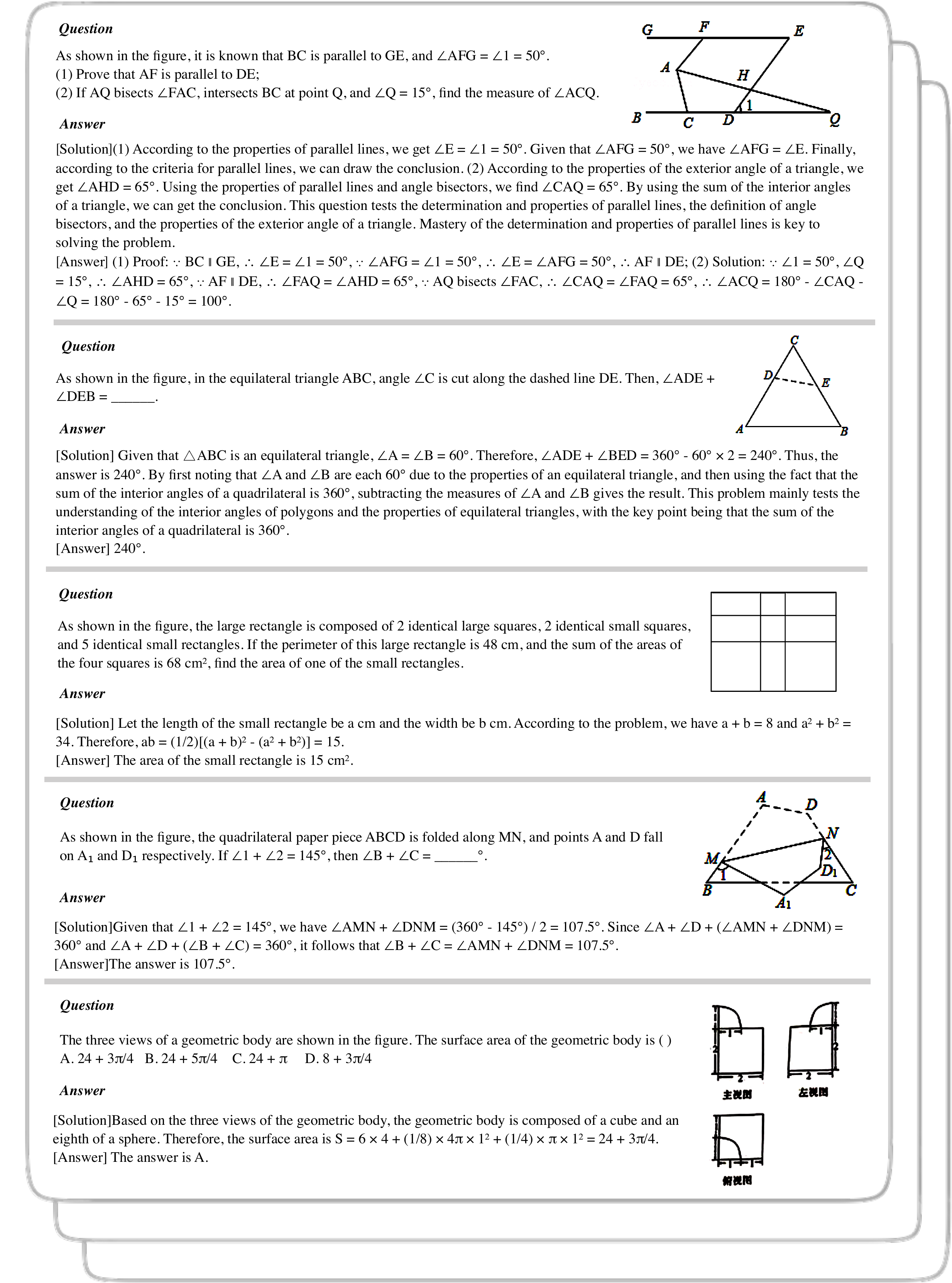}
    \vspace{-2mm}
    \caption{Cases of geometric problems in our \data dataset.}
    \label{fig:dataset_geometry}
    \vspace{-3mm}
\end{figure}

\begin{figure}[hbpt]
    \centering
    \includegraphics[width=1.0\textwidth]{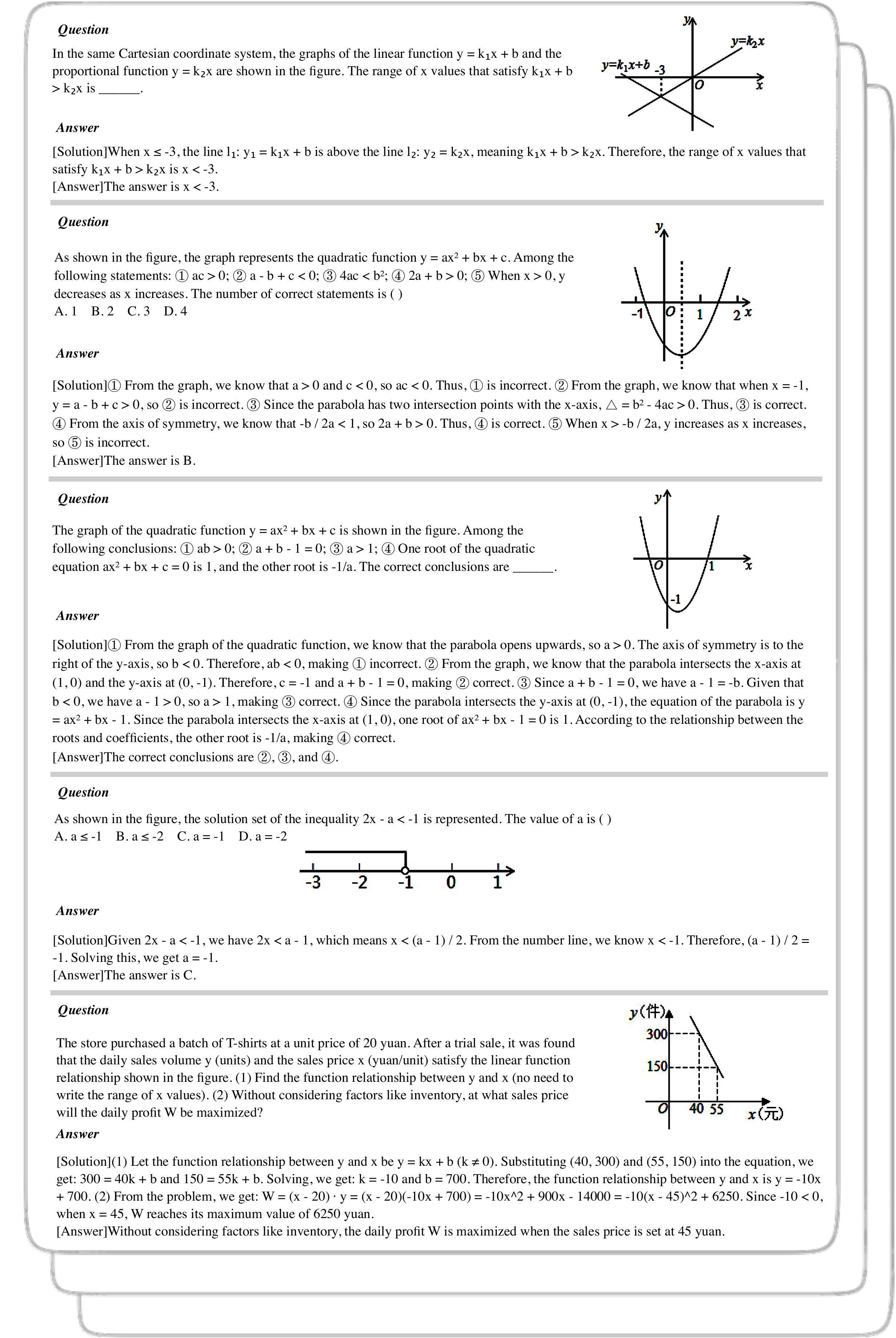}
    \vspace{-2mm}
    \caption{Cases of algebraic problems in our \data dataset.}
    \label{fig:dataset_algebra}
    \vspace{-3mm}
\end{figure}

\begin{figure}[hbpt]
    \centering
    \includegraphics[width=1.0\textwidth]{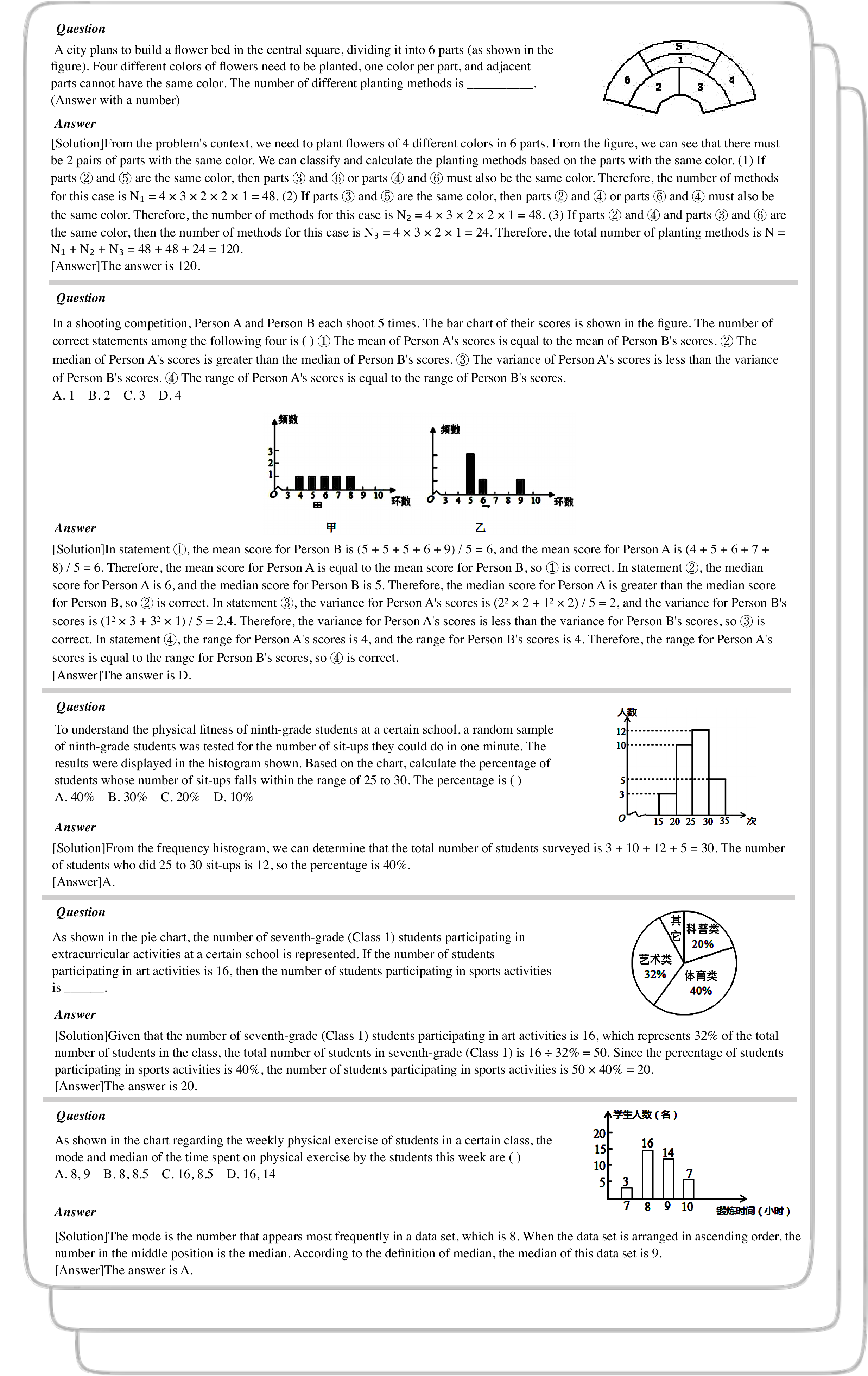}
    \vspace{-2mm}
    \caption{Cases of statistical problems in our \data dataset.}
    \label{fig:dataset_statistics}
    \vspace{-3mm}
\end{figure}

\begin{figure}[hbpt]
    \centering
    \includegraphics[width=1.0\textwidth]{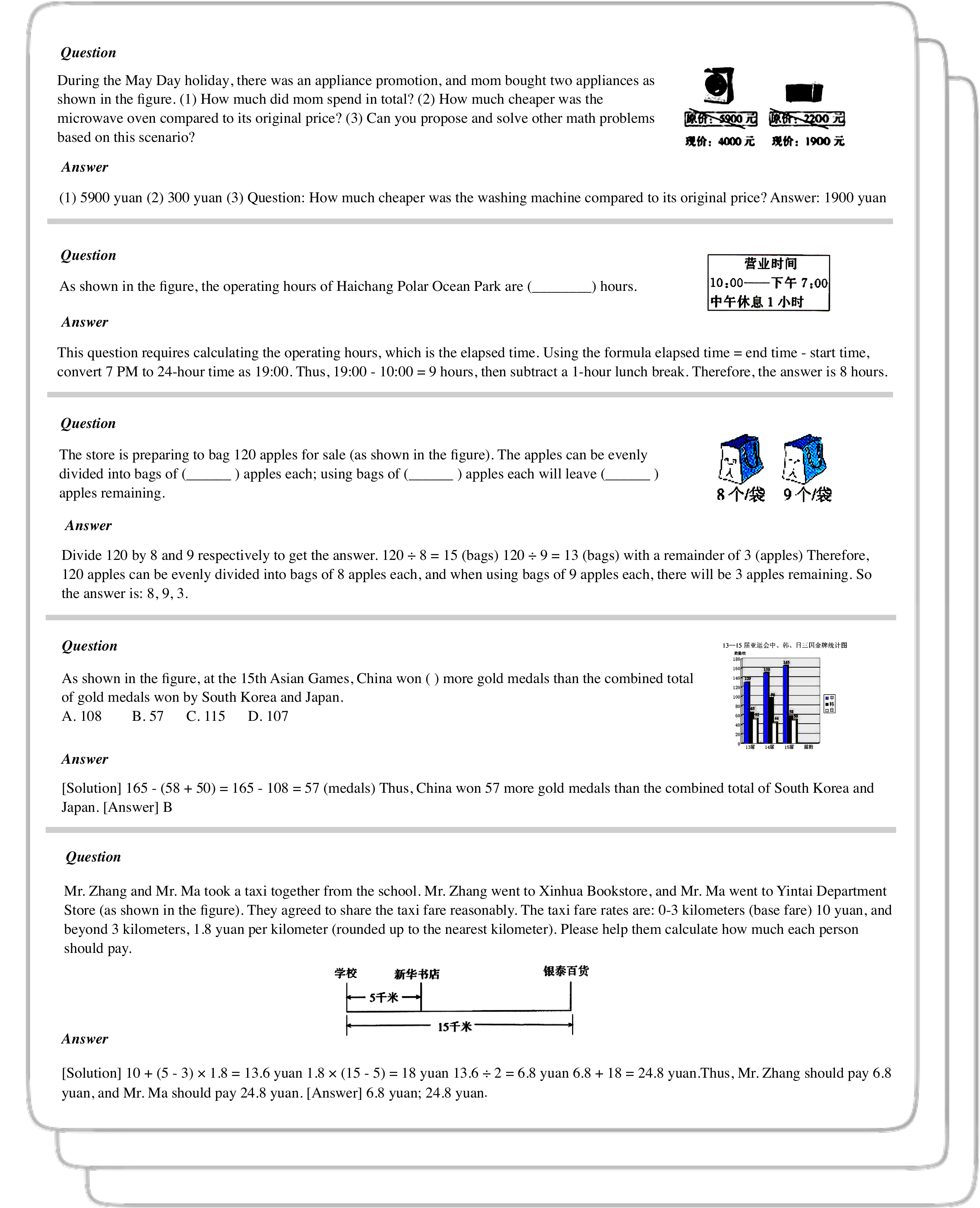}
    \vspace{-2mm}
    \caption{Cases of word problems in our \data dataset.}
    \label{fig:dataset_word_problems}
    \vspace{-3mm}
\end{figure}

In this section, we provide a detailed overview of specific cases from our constructed \data dataset. These cases demonstrate the variety of mathematical disciplines covered by \data, including arithmetic, geometry, algebra, statistics, and word problems. Figure~\ref{fig:dataset_arithmetic}, Figure~\ref{fig:dataset_geometry}, Figure~\ref{fig:dataset_algebra}, Figure~\ref{fig:dataset_statistics}, and Figure~\ref{fig:dataset_word_problems}  depict the types of problems that \model is designed to tackle in each respective category. Each of these categories is critical for assessing the comprehensive mathematical capabilities of \model. By tackling a wide range of problems, \model demonstrates its versatility and robustness in addressing diverse mathematical tasks. 

\section{Backbone Models}\label{appendix: backbone_models}

We utilize the following multi-modal large language models as our backbone models for conducting Specialized Fine-Tuning (SFT) on the constructed \data. The detailed description of each backbone model can be represented as follows:

\begin{itemize}
    \item \textbf{GLM-4V-9B} is a bilingual (Chinese and English) multi-modal large language model, developed collaboratively by Zhipu.AI and Tsinghua University. It is built upon the foundational architecture of GLM-4-9B, enhancing its capabilities to handle complex multi-modal interactions. GLM-4V-9B takes a high resolution of 1120 * 1120 images as visual inputs.  In comprehensive evaluations that test various capabilities including combined language skills, perceptual reasoning, text recognition, and chart understanding, GLM-4V-9B consistently outperforms competitors such as GPT-4-turbo-2024-04-09, Gemini 1.0 Pro, Qwen-VL-Max, and Claude 3 Opus, demonstrating its superior performance across multiple modalities.

    \item \textbf{CogVLM2} is a series of open-source multi-modal large language models derived from Meta-Llama-3-8B-Instruct, developed by Zhipu.AI and Tsinghua University. This series contains two models: cogvlm-llama3-chat-19B and cogvlm2-llama3-chinese-chat-19B. The former is a monolingual language model focused on English and the latter is a bilingual model supporting both English and Chinese. CogVLM2 is designed to handle extended content lengths up to 8K and accepts high-resolution images up to 1344 * 1344. Here, we choose cogvlm2-llama3-chinese-chat-19B as our backbone to pre-train our \model-19B. 

    \item \textbf{CogVLM-32B} is a close-source multi-modal large language model, developed by Zhipu.AI and Tsinghua University. It is based on the GLM-32B architecture and is optimized for handling complex multi-modal tasks. CogVLM-32B is engineered to process visual inputs at a high resolution of 1120 * 1120, enabling detailed image analysis and enhanced interaction with visual data.

\end{itemize}

\section{Descriptions of VQA Datasets}\label{appendix: vqa_datasets_details}
Here, we provide a detailed description of collected visual question answering datasets (VQA) datasets. Table~\ref{tab:vqa_datasets_details} demonstrates details 19 different VQA datasets, including task types and visual context.

\begin{table}[hbpt]
    \centering
    \renewcommand{\arraystretch}{1.15}
    \resizebox{0.7\textwidth}{!}{%
    \begin{tabular}{ccc}  
    \toprule
     Dataset    & Task & Visual Context  \\
    \midrule
    DocVQA & Figure Question Answering (FQA) & Document Image \\
    DVQA & Figure Question Answering (FQA) & Bar Chart \\
    FigureQA & Figure Question Answering (FQA) & Charts and Plots \\
    PlotQA & Figure Question Answering (FQA) & Bar, Line, Scatter \\
    MapQA & Figure Question Answering (FQA) & Map Chart \\
    \midrule
    IconQA &  Math Word Problem (MWP) & Abstract Scene \\
    TabMWP & Math Word Problem (MWP) & Table \\
    CLEVR-Math & Math Word Problem (MWP) & Synthetic Scene \\
    \midrule
    TQA & Textbook Question Answering (TQA) & Scientific Figure \\
    AI2D & Textbook Question Answering (TQA) & Scientific Figure \\
    ScienceQA & Textbook Question Answering (TQA) & Scientific Figure \\
    \midrule
    A-OKVQA & Visual Question Answering (VQA) & Natural Image\\
    VQA2.0 & Visual Question Answering (VQA) & Natural Image \\
    PMC-VQA & Visual Question Answering (VQA) & Medical Image\\
    VizWiz & Visual Question Answering (VQA) & Natural Image \\
    Super-CLEVR & Visual Question Answering (VQA) & Synthetic Scene \\
    VQA-AS & Visual Question Answering (VQA) & Abstract Scene\\
    VQA-RAD & Visual Question Answering (VQA) & Medical Image\\
    TextVQA & Visual Question Answering (VQA) & Natural Image \\
    \bottomrule
    \end{tabular}} 
    \vspace{0.2cm}
    \caption{Summary of VQA datasets.}
    \label{tab:vqa_datasets_details}
\end{table}

\section{Implementation Details}\label{appendix: finetuning_setup}

We provide a detailed overview of the Specialized Fine-Tuning (SFT) process applied to our \model. The specific hyperparameters used during this process are outlined in Table~\ref{tab:implementation_details}.

\begin{table}[hbpt]
    \centering
    \renewcommand{\arraystretch}{1.15}
    \resizebox{\textwidth}{!}{%
    \begin{tabular}{c|ccc}  
    \toprule
     parameters    & \model-9B & \model-19B & \model-32B  \\
    \midrule
     Total steps & 35,000 & 35,000 & 35,000 \\
     Global Batch Size & 128 & 128 & 128   \\
     Learning Rate & $1e^{-5}$  & $1e^{-5}$ & $1e^{-5}$  \\
     Learning Rate Schedule & cosine decay  & cosine decay  & cosine decay  \\
     Warmup Ratio & 0.01 & 0.01 & 0.01  \\
     Weight Decay & $5e^{-2}$   & $5e^{-2}$   & $5e^{-2}$   \\
     Optimizer & AdamW   & AdamW  & AdamW \\
     Input Resolution  & 1120 * 1120 & 1344 * 1344 & 1120 * 1120  \\
     Image Length  & 1600 & 2304 & 1600 \\
    \bottomrule
    \end{tabular}} 
    \vspace{0.2cm}
    \caption{The detailed setup of the SFT procedures.}
    \label{tab:implementation_details}
\end{table}

\section{The detailed description of Benchmark Datasets}\label{appendix: benchmark_dataset}

In this section, we provide an in-depth description of the benchmark datasets used to evaluate the performance of \model. These benchmark datasets have been carefully curated to test the MLLMs' capabilities. The detailed description of benchmark datasets is provides as follows.

\begin{itemize}
    \item \textbf{MathVista}
    
    MathVista is a comprehensive benchmark dataset designed to rigorously evaluate the mathmetical reasoning capabilities of language models (LMs), especially in varied visual contexts. 

    This dataset offers a comprehensive evaluation benchmark designed to integrate mathematical reasoning with visual understanding, focusing on five primary tasks: figure question answering (FQA); geometry problem solving (GPS); math word problem (MWP); textbook question answering (TQA); and visual question answering (VQA).
    
    \item \textbf{MathVista-GPS}

    MathVista-GPS, a subset of the MathVista Dataset, specifically focuses on the domain of geometry problem solving. The questions in this subset range from basic shape recognition to more advanced problems involving theorems, calculations and reasoning.
    
    \item \textbf{MathVerse}
    
    MathVerse is designed to provide a fair and comprehensive assessment of MLLMs' capabilities in visual mathematics. The benchmark comprises 2,612 high-quality, multi-subject math problems, each featuring diagrams and converted into six different versions by human annotators. These versions offer varying levels of multi-modal information, allowing for a thorough evaluation of MLLMs' understanding of visual diagrams.

    \item \textbf{MATH-Vision} 
    
    The MATH-Vision (MATH-V) dataset comprises 3,040 high-quality mathematical problems, each featuring a visual context and sourced from 19 real math competitions. This extensive and diverse collection allows for a comprehensive evaluation of LMMs' ability to interpret and reason with visual information in mathematical contexts.

    \item \textbf{MMMU}

    The Massive Multi-discipline Multi-modal Understanding and Reasoning (MMMU) benchmark encompasses 11.5K questions across six disciplines, including Art, Business, Health \& Medicine, Science, Humanities \& Social Science, and Tech \& Engineering. The tasks in MMMU challenge models to perform sophisticated multi-modal analysis and apply domain-specific knowledge, demanding a higher level capability in comprehension and integration.
    
\end{itemize}

\section{Detailed Experimental Results on Public Benchmark Datasets}\label{appendix:detailed_public}

\vpara{Results on the testmini subset of MathVista.} To comprehensively evaluate the performance of \model across various task types featured in the MathVista dataset, we systematically evaluate it on the testmini subset. This subset has been carefully selected to represent a diverse range of mathematical problem types, ensuring a robust assessment of our model's capabilities. Table~\ref{tab:mathvista_test_result} shows the evaluation results on the testmini subset of MathVista across various task types. Notably, \model-19B and \model-32B surpass human performance in overall accuracy, highlighting the advanced capabilities of these models in handling complex mathematical problems. In particular, \model excels significantly in geometry problem solving (GPS) and geometry reasoning (GEO), demonstrating its superior proficiency in mathematical reasoning.  

\begin{table*}[h]
\centering
 \small
 \renewcommand\arraystretch{1.15} 
\begin{adjustbox}{width=\linewidth}
    \begin{tabular}{l|c|c|ccccc|ccccccc}
    \toprule
    Model & Input & ALL & FQA & GPS& MWP & TQA & VQA & ALG & ARI & GEO & LOG & NUM & SCI& STA  \\ 
    \midrule
    Human Performance & $Q$, $I$ & 60.30 & 59.70 & 48.40 & \cellcolor{red!25}{73.00} & \cellcolor{red!25}{63.20} & \cellcolor{red!25}{55.90} & 50.90 & 59.20 & 51.40 & \cellcolor{red!25}{40.70} & 53.80 & \cellcolor{red!25}{64.90} & 63.90  \\
    \midrule
    2-shot CoT GPT-4 & $Q$, $I_c$, $I_t$ & 30.50 & 27.21 & 35.91 & 21.30 & 43.13 & 28.17 & 35.72 & 25.17 & 35.80 & 24.74 & 15.41 & 47.28 & 31.29 \\
    2-shot PoT GPT-4 & $Q$, $I_c$, $I_t$ & 31.74 & 27.58 & 37.35 & 23.87 & 43.00 & 30.27 & 37.15 & 27.93 & 37.48 & 22.68 & 15.83 & 44.47 & 31.87 \\
    GPT-4V & $Q$, $I$  & 49.90	& 43.10	& 50.50 &	57.50	& 65.20	& 38.00	& 53.00	& 49.00 & 51.00 & 21.60	& 20.10	& 63.10 & 55.80 \\
    \midrule
    LLaVA-LLaMA-2-13B & $Q$, $I$ & 25.40 & 22.86 & 24.57 & 18.15 & 35.82 & 29.69 & 26.93 & 22.47 & 24.45 & 19.07 & 19.05 & 34.71 & 21.61 \\
    \midrule 
    \textbf{\model-9B} & $Q$, $I$ & 52.20 & 46.10 & 64.42 & 58.60 & 55.70 & 37.43 & 59.79 & 43.91 & 62.34 & 10.81 & 37.50 & 54.10 & 54.82\\
    \textbf{\model-19B} & $Q$, $I$ & 61.10 & 59.85 & \textbf{65.38} & 68.28 & 53.80 & \textbf{55.31} & 59.79 & 59.21 & \textbf{63.18} & \textbf{18.92} & \textbf{59.03} & 53.28 & 68.44\\
    \textbf{\model-32B} & $Q$, $I$ & \textbf{62.40} & \textbf{62.83} & 62.02 & \textbf{69.35} & \textbf{62.03} & 54.19 & \textbf{60.50} & \textbf{60.62} & 61.92 & 16.22 & 52.08 & \textbf{60.66} & \textbf{72.09} \\
    \bottomrule
    \end{tabular}
    \end{adjustbox}
    \caption{\textbf{Accuracy scores on the \textit{testmini} subset of MathVista.} Input: $Q$: question, $I$: image, $I_c$: image caption, $I_t$: OCR texts detected from the image. ALL: overall accuracy. Task types: FQA: figure question answering, GPS: geometry problem solving, MWP: math word problem, TQA: textbook question answering, VQA: visual question answering. Mathematical reasoning types: ALG: algebraic reasoning, ARI: arithmetic reasoning, GEO: geometry reasoning, LOG: logical reasoning, NUM: numeric common sense, SCI: scientific reasoning, STA: statistical reasoning. The highest accuracy among all baseline MLLMs is marked in red, while the highest accuracy among various variants of \model is marked bold.}
\label{tab:mathvista_test_result}
\end{table*}

\vpara{Results on the testmini set of MathVerse.} To thoroughly evaluate the performance of \model across 12 detailed subjects within the MathVerse dataset, we conduct comprehensive experiments and report the results in Table~\ref{tab:mathverse_test_result}. This analysis delves into the model's ability to address a broad spectrum of mathematical challenges, ranging from geometry to functions. As shown in Table~\ref{tab:mathverse_test_result}, \model surpasses all open-source MLLMs and most close-source MLLMs. However, it still falls short by 14\% compared to the performance of GPT-4V. In some subjects such as Angle, Analytic, and Property, \model achieves better performance compared the advanced GPT-4V. For example, \model-32B shows remarkable performance in plane geometry, particularly in handling angle-related problems, where it achieves a 60.1\% accuracy, showcasing its strong geometric reasoning capabilities.

\begin{table*}[h]
\small
\centering
\begin{adjustbox}{width=\linewidth}
    \begin{tabular}{l|c|cccccc|cccc|ccccc}
    \toprule
    \multirow{3}*{\makecell*[l]{\large Model}}    &\multirow{3}*{\makecell*[c]{All}}
    &\multicolumn{6}{c|}{\makecell*[c]{\shortstack{Plane Geometry}}} 
    &\multicolumn{4}{c|}{\makecell*[c]{\shortstack{Solid Geometry}}}
    &\multicolumn{5}{c}{\makecell*[c]{\shortstack{Functions}}}\\
    \cmidrule{3-17}
    & &\makecell*[c]{All} &\makecell*[c]{Len} &\makecell*[c]{Area} &\makecell*[c]{Angle} &\makecell*[c]{Anal} &\makecell*[c]{Apply} &\makecell*[c]{All} &\makecell*[c]{Len} &\makecell*[c]{Area} &\makecell*[c]{Vol} &\makecell*[c]{All} &\makecell*[c]{Coord} &\makecell*[c]{Prop} &\makecell*[c]{Exp} &\makecell*[c]{Apply} \\
    \midrule
    \multicolumn{17}{c}{\textit{Closed-source MLLMs}}\\
    \cmidrule{1-17}
    Qwen-VL-Plus&21.3&17.3&19.1&16.4&16.1&23.6&13.2&24.8&18.1&18.7&33.4&31.3&52.5&25.1&10.8&50.3\\
    Gemini-Pro & 35.3 & 33.0 & 32.2 & 42.6  & 28.4 & 30.2 & 32.3 & 33.4 & 35.0 & 29.3 & 36.1  & 28.3 & 25.7  & 26.6& 10.8  & 51.3 \\
    Qwen-VL-Max & 37.2 & 38.4 & 41.7 & 46.4  & 32.6 & 40.6 & 38.7 & 33.7 & 25.4 & 28.3 & 42.6  & 38.4 & 43.7  & 35.5& 13.6  & 61.0 \\
    GPT-4V & \cellcolor{red!25}{54.4} & \cellcolor{red!25}{56.9} & \cellcolor{red!25}{60.8} & \cellcolor{red!25}{63.4}  & \cellcolor{red!25}{52.6} & {48.5} & \cellcolor{red!25}{60.9} & {50.2} & {54.8} & {39.9} & \cellcolor{red!25}{56.8} & \cellcolor{red!25}{52.8} & \cellcolor{red!25}{72.3}  & {47.1} & {30.9}  & \cellcolor{red!25}{70.1} \\
    \cmidrule{1-17}
    \multicolumn{17}{c}{\textit{Open-source MLLMs}}\\
    \cmidrule{1-17}
    LLaMA-Adapter V2&5.8 & 5.9&4.0&5.9&6.6&13.4&3.3&4.6&5.3&3.1&5.7&6.2&6.7&6.1&4.5&7.9\\
    ImageBind-LLM & 10.0& 9.7&12.1&9.9&9.2&10.2&4.8&4.6&4.9&3.5&5.3&14.9&12.3&13.8&4.6&25.9 \\
    mPLUG-Owl2 &10.3& 7.7&8.2&6.0&5.7&12.4&10.6&11.0&9.2&6.7&15.7&17.4&22.8&18.6&5.3&22.2\\
    MiniGPT-v2 &10.9&11.6&10.0&9.8&14.3&9.1&11.8&1.7&2.2&1.6&0.5&11.2&4.2&15.7&4.0&21.1\\
    LLaVA-1.5 & 12.7 & 11.8 & 13.1 & 15.1 & 9.7 & 9.4 & 13.2 & 10.6 & 12.1 & 8.7 & 11.6  & 14.8 & 18.8  & 12.7& 9.5  & 23.7\\
    SPHINX-Plus &14.0&14.4&14.2&10.5&14.1&16.5&16.8&7.0&7.2&6.1&7.6&17.9&11.1&19.1&6.3&27.7 \\
    G-LLaVA & 15.7 & 20.2 & 17.3&13.6&{26.5}&5.9&23.1&5.0&10.3&4.4&3.1&9.2&9.1&9.1&1.3&15.5\\
    LLaVA-NeXT & 17.2 & 15.9 & 14.8 & 13.1  & 16.3 & 17.7 & 17.8 & 19.6 & 33.3 & 11.7 & 12.6  & 23.1 & {24.5}  & 23.4 & 8.0  & 33.1\\
    ShareGPT4V &17.4&16.9&16.2&17.9&16.9&12.2&21.1&15.0&13.6&10.9&19.7&20.2&19.9&22.2&8.4&25.8\\
    SPHINX-MoE & 22.8 & 24.5 & 26.3 & 28.4  & 21.1 & {26.6} & {24.4} & 15.8 & 9.4 & 10.7 & {26.3}  & 19.5 & 23.5  & 19.3& 9.2  & 30.3\\
    InternLM-XC2 & {25.9} & {26.2} & {27.1} & {29.7}  & 20.6 & 18.5 & 22.2 & {20.1} & {34.5} & {14.1} & 25.2  & {23.7} & 24.4  & {24.9} & {10.6}  & {36.3}\\
    \cmidrule{1-17}
    \multicolumn{17}{c}{\textit{\model}}\\
    \cmidrule{1-17}
    \textbf{\model-9B} & 44.2 & 45.3 & \textbf{43.7} & 48.9 & 41.5 & \textbf{53.5} & 52.2 & 42.0 & 54.2 & 50.0 & 29.4 & 42.1 & 25.0 & 42.3 & \textbf{43.8} & 47.5 \\
    \textbf{\model-19B} & 42.5 & 41.8 & 34.8 & \textbf{55.3} & 38.9 & 46.5 & \textbf{53.6} & \textbf{51.3} & \textbf{66.7} & \textbf{52.3} & \textbf{43.1} & 38.4 & 18.8 & 38.0 & 28.1 & 55.0 \\
    \textbf{\model-32B} & \textbf{49.2} & \textbf{49.0} & 42.4 & 59.6 & \textbf{51.3} & 48.8 & 50.7 & 45.4 & 62.5 & 40.9 & 41.2 & \textbf{52.8} & \textbf{43.8} & \textbf{59.2} & 40.6 & \textbf{55.0} \\
    \bottomrule
    \end{tabular}
\end{adjustbox}
\caption{\textbf{Mathematical Evaluation on Different Subjects and Subfields in MathVerse's \textit{testmini} Set.} Len: Length; Anal: Analytic; Apply: Applied; Vol: Volume; Coord: Coordinate; Prop: Property; Exp: Expressio. The highest accuracy among all baseline MLLMs is marked in red, while the highest accuracy among various variants of \model is marked bold.}
\label{tab:mathverse_test_result}
\end{table*}

\vpara{Results on Math-Vision datasets.} To effectively assess \model's ability across diverse subjects and difficulty levels within the Math-Vision dataset, we conduct a series of detailed evaluation experiments and report results in Table~\ref{tab:mathvision_test_result}. Specifically, GPT-4V leads the close-source models with an overall accuracy of 22.76\%, yet it remains significantly below the human performance benchmark of 75.66\%. \model shows competitive performance across a variety of mathematical disciplines compared to most of close-source MLLMs, with \model-32B achieving the overall accuracy of 26.5\%, closely approaching that of GPT-4V. Notably, \model-32B excels in solid geometry with a accuracy of 29.1\%, significantly outperforming the accuracy of 23.8\% on GPT-4V. This superior performance in solid geometry highlights \model-32B's advanced spatial reasoning and geometric processing capabilities, which are essential for tackling complex three-dimensional problems.

\begin{table*}[t!]
\centering
\resizebox{\textwidth}{!}{%
\begin{tabular}{l|c|cccccccccccccccc}
\toprule
\multicolumn{18}{c}{Human Performance}\\
\midrule
Model & Overall & Alg & AnaG & Ari & CombG & Comb & Cnt & DescG & GrphT & Log & Angle & Area & Len & SolG & Stat & Topo & TransG \\
\toprule
Human (testmini) & 75.66 & 57.9 & 79.0 & 100.0 & 100.0 & 47.4 & 94.7 & 89.5 & 63.2 & 63.2 & 36.8 & 52.6 & 73.7 & 89.5 & 89.5 & 100.0 & 73.7 \\
\midrule
\multicolumn{18}{c}{Open-source MLLMs}\\
\midrule
LLaVA-v1.5-7B & 8.52 & 7.0 & 7.1 & 10.7 & 7.1 & 4.8 & 10.5 & 7.7 & 10.0 & 9.2 & 15.6 & 10.2 & 9.8 & 5.3 & 8.6 & 4.4 & 4.8 \\
SPHINX (V2) & 9.70 & 6.7 & 7.1 & 12.9 & 7.5 & 7.7 & 6.0 & 9.6 & 16.7 & 10.1 & 11.0 & 11.8 & 12.5 & 8.2 & 8.6 & 8.7 & 6.0 \\
ShareGPT4V-7B & 10.53 & 5.5 & 3.6 & 12.9 & 10.1 & 4.8 & 7.5 & 11.5 & 14.4 & 10.9 & 16.2 & 11.8 & 12.3 & 9.8 & 15.5 & 17.4 & 11.3 \\
LLaVA-v1.5-13B & 11.12 & 7.0 & 14.3 & 14.3 & 9.1 & 6.6 & 6.0 & 13.5 & 5.6 & 13.5 & 10.4 & 12.6 & 14.7 & 11.5 & 13.8 & 13.0 & 10.7 \\
ShareGPT4V-13B & 11.88 & 7.5 & 15.5 & 16.4 & 10.7 & 8.9 & 9.0 & 11.5 & 8.9 & 7.6 & 11.6 & 13.0 & 17.4 & 10.3 & 8.6 & 8.7 & 12.5 \\
SPHINX-MoE & 14.18 & 7.8 & 17.9 & 14.3 & 15.6 & 9.5 & 11.9 & 12.5 & 15.6 & 12.6 & 16.2 & 15.6 & 17.8 & 13.5 & 12.1 & 8.7 & 16.1 \\
InternLM-XComposer2-VL & 14.54 & 9.3 & 15.5 & 12.1 & 15.3 & 11.3 & 10.5 & 14.4 & {22.2} & {19.3} & {19.7} & 15.6 & 15.0 & 11.9 & 15.5 & {26.1} & 15.5 \\
\midrule
\multicolumn{18}{c}{Closed-source MLLMs}\\
\midrule
Qwen-VL-Plus & 10.72 & 11.3 & 17.9 & 14.3 & 12.7 & 4.8 & 10.5 & 15.4 & 8.9 & 14.3 & 11.6 & 6.4 & 10.0 & {14.3} & 6.9 & 8.7 & 11.31 \\
Qwen-VL-Max & 15.59 & 10.7 & {19.1} & 20.0 & 16.9 & {12.5} & {17.9} & 16.4 & 12.2 & {21.0} & 13.3 & 14.2 & 19.8 & 11.5 & {20.7} & 13.0 & 17.3 \\
Gemini Pro & {17.66} & {15.1} & 10.7 & {20.7} & {20.1} & 11.9 & 7.5 & {20.2} & {21.1} & 16.8 & 19.1 & {19.0} & {20.0} & {14.3} & 13.8 & 17.4 & {20.8} \\
GPT4V & \cellcolor{red!25}{22.76} & \cellcolor{red!25}{27.3} & \cellcolor{red!25}{32.1} & \cellcolor{red!25}{35.7} & {21.1} & {16.7} & {13.4} & {22.1} & 14.4 & 16.8 & {22.0} & {22.2} & {20.9} & {23.8} & {24.1} & \cellcolor{red!25}{21.7} & {25.6} \\
\midrule
\multicolumn{18}{c}{\model}\\
\midrule
\textbf{\model-9B}  & 19.2 & 18.6 & 20.2 & 19.3 & 15.3 & 18.5 & 20.9 & 26.0 & 18.9 & 15.1 & 23.1 & 20.4 & 18.3 & 23.8 & 19.0 & 17.4 & 14.3  \\
\textbf{\model-19B} & 21.6 & 22.0 & \textbf{29.8} & 23.6 & 22.4 & 18.5 & \textbf{25.4} & 25.0 & 17.8 & 16.0 & 20.2 & 22.0 & 20.3 & 21.3 & 20.7 & \textbf{30.4} & 23.2\\
\textbf{\model-32B} & \textbf{26.5} & \textbf{22.9} & 20.2 & \textbf{24.3} & \textbf{23.1} & \textbf{28.0} & 20.9 & \textbf{34.6} & \textbf{27.8} & \textbf{23.5} & \textbf{31.2} & \textbf{26.8} & \textbf{30.1} & \textbf{29.1} & \textbf{22.4} & 17.4 & \textbf{26.2} \\
\bottomrule
\end{tabular}%
}
\caption{\textbf{Comparison of model performances across various mathematical subjects.} Subjects: Alg: algebra, AnaG: analytic geometry, Ari: arithmetic, CombG: combinatorial geometry, Comb: combinatorics, Cnt: counting, DescG: descriptive geometry, GrphT: graph theory, Log: logic, Angle: metric geometry - angle, Area: metric geometry - area, Len: metric geometry - length, SolG: solid geometry, Stat: statistics, Topo: topology, TransG: transformation geometry. The highest accuracy among all baseline MLLMs is marked in red, while the highest accuracy among various variants of \model is marked bold.}
\label{tab:mathvision_test_result}
\vspace{-0.2cm}
\end{table*}

\section{Error Cases}\label{appendix: error_cases}

\begin{figure}[hbpt]
    \centering
    \includegraphics[width=1.0\textwidth]{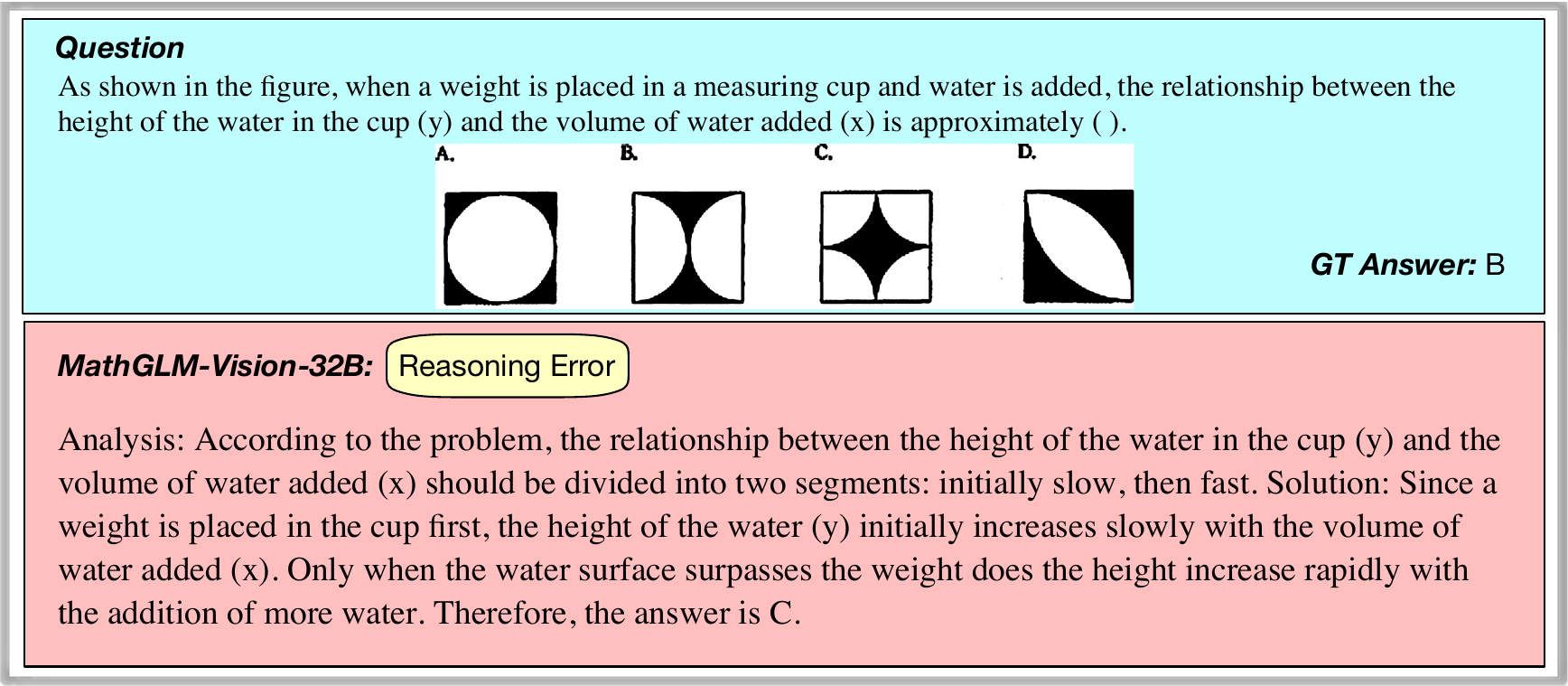}
    \vspace{-2mm}
    \caption{\textbf{An example of reasoning error. } \model-32B mistakenly believed that the height of the water would initially increase rapidly and then slowly, indicating a reasoning error.}
    \label{fig:case_error_1}
    \vspace{-3mm}
\end{figure}
\begin{figure}[hbpt]
    \centering
    \includegraphics[width=1.0\textwidth]{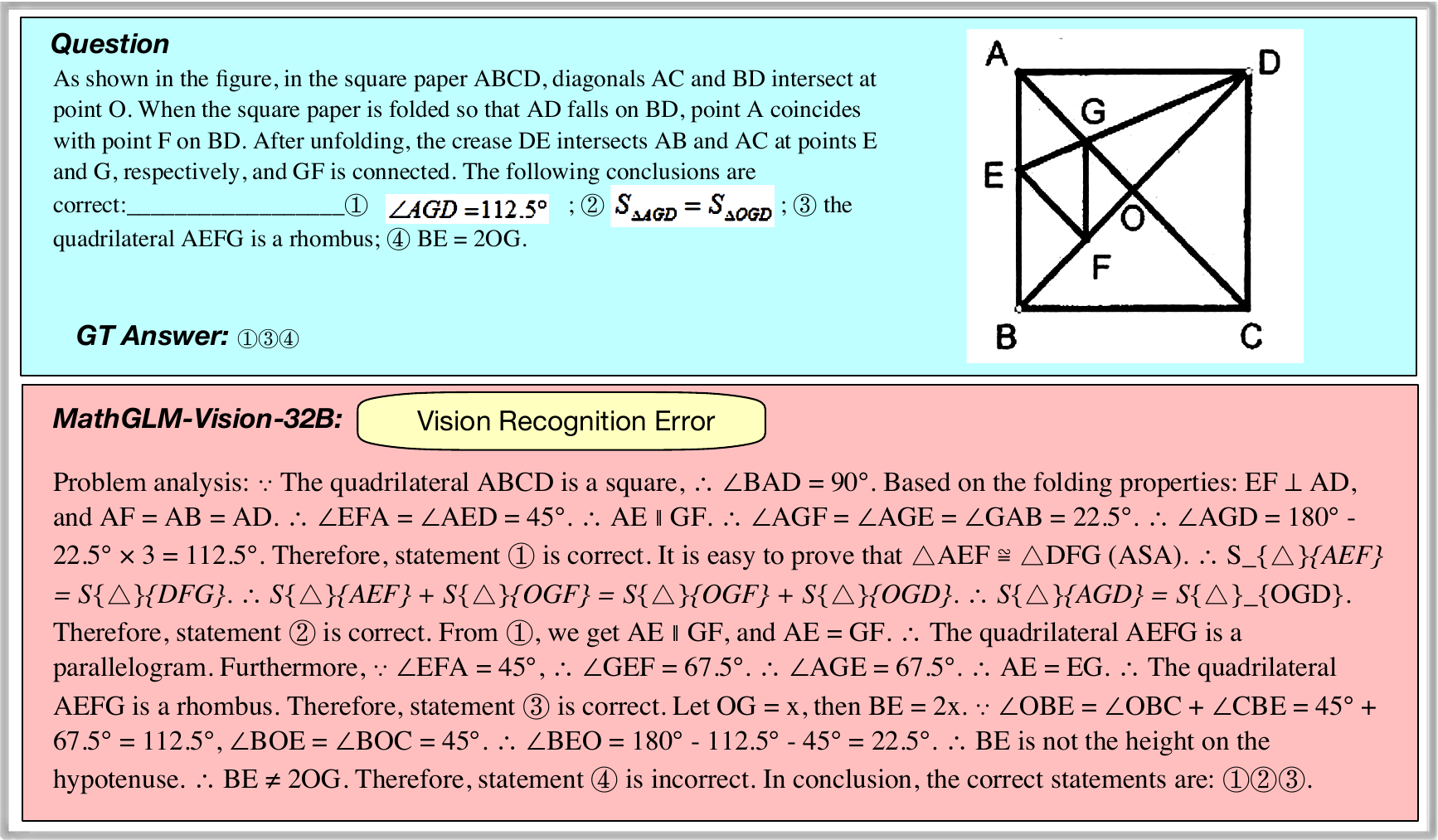}
    \vspace{-2mm}
    \caption{\textbf{An example of vision recognition error. } \model-32B incorrectly interpreted the geometric properties of the diagram, leading to a vision recognition error.}
    \label{fig:case_error_2}
    \vspace{-3mm}
\end{figure}
\begin{figure}[hbpt]
    \centering
    \includegraphics[width=1.0\textwidth]{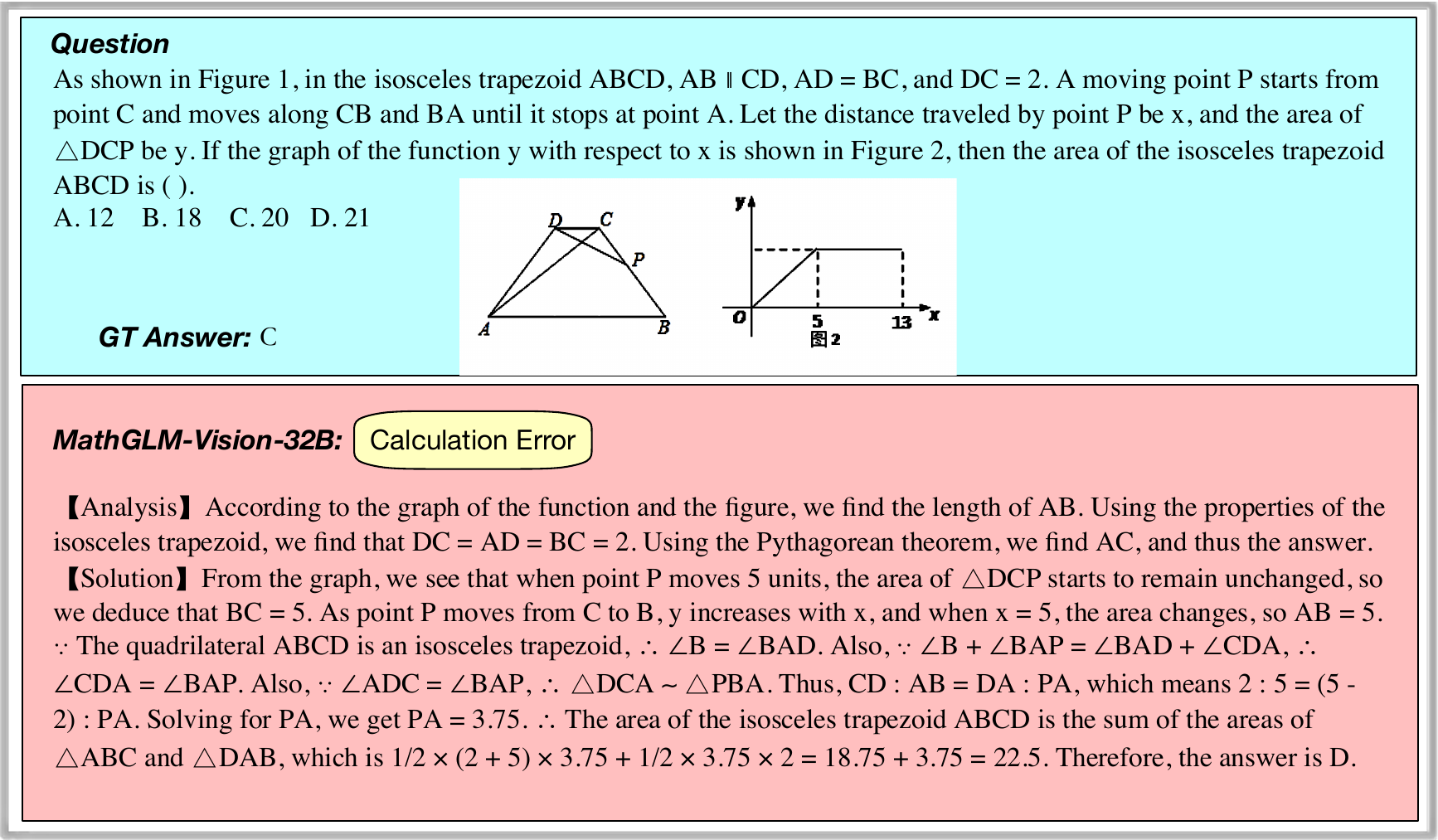}
    \vspace{-2mm}
    \caption{\textbf{An example of calculation error. } \model-32B made an error in calculating the lengths and areas in the isosceles trapezoid, leading to an incorrect solution.}
    \label{fig:case_error_3}
    \vspace{-3mm}
\end{figure}
\begin{figure}[hbpt]
    \centering
    \includegraphics[width=1.0\textwidth]{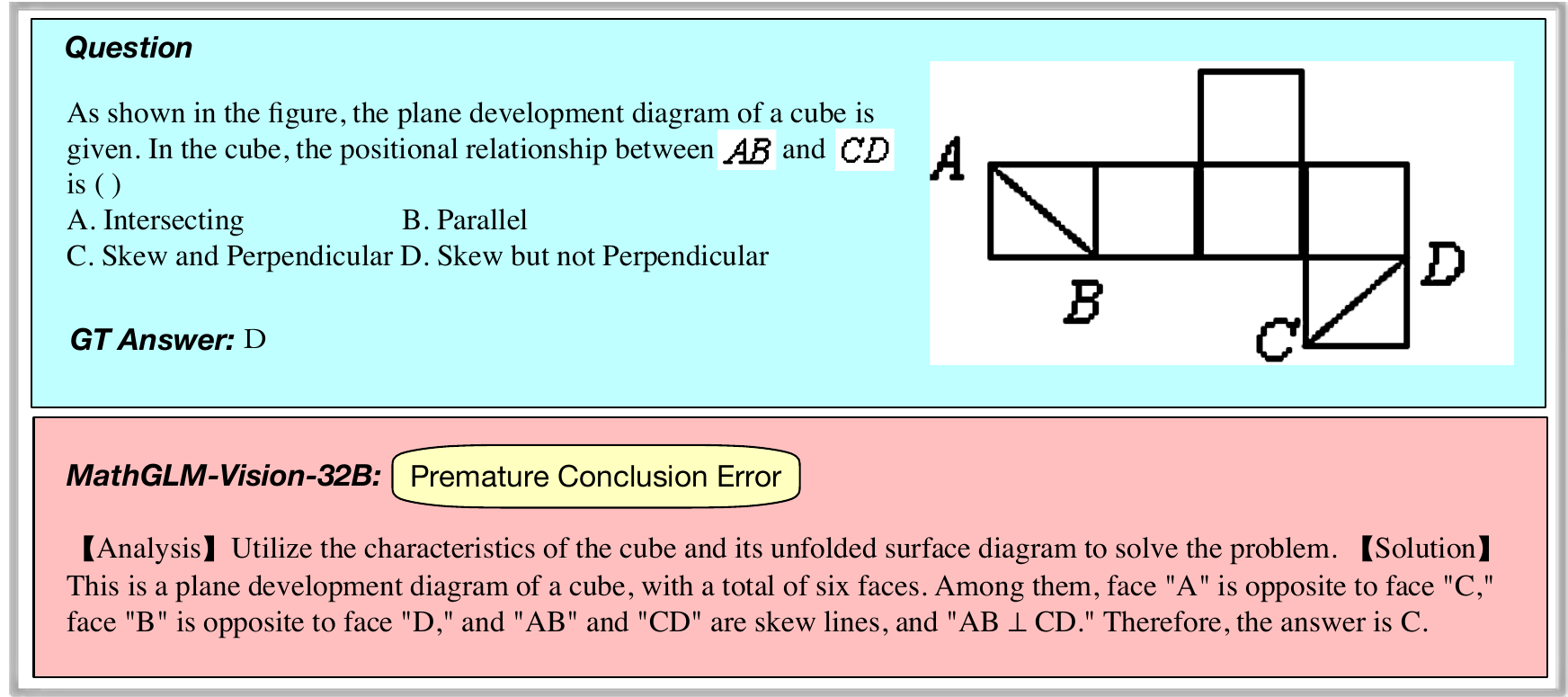}
    \vspace{-2mm}
    \caption{\textbf{An example of premature conclusion error. } \model-32B prematurely concluded that AB is perpendicular to CD without proper reasoning, leading to a premature conclusion error.}
    \label{fig:case_error_4}
    \vspace{-3mm}
\end{figure}

Figure ~\ref{fig:case_error_1}, Figure~\ref{fig:case_error_2}, Figure~\ref{fig:case_error_3}, and Figure~\ref{fig:case_error_4} show examples of errors made by \model-32B on the \data-test dataset. Each figure highlights a specific type of error, providing valuable insights into the model's limitations and areas for improvement.

\section{Case Study}

\begin{figure}[hbpt]
    \centering
    \includegraphics[width=1.0\textwidth]{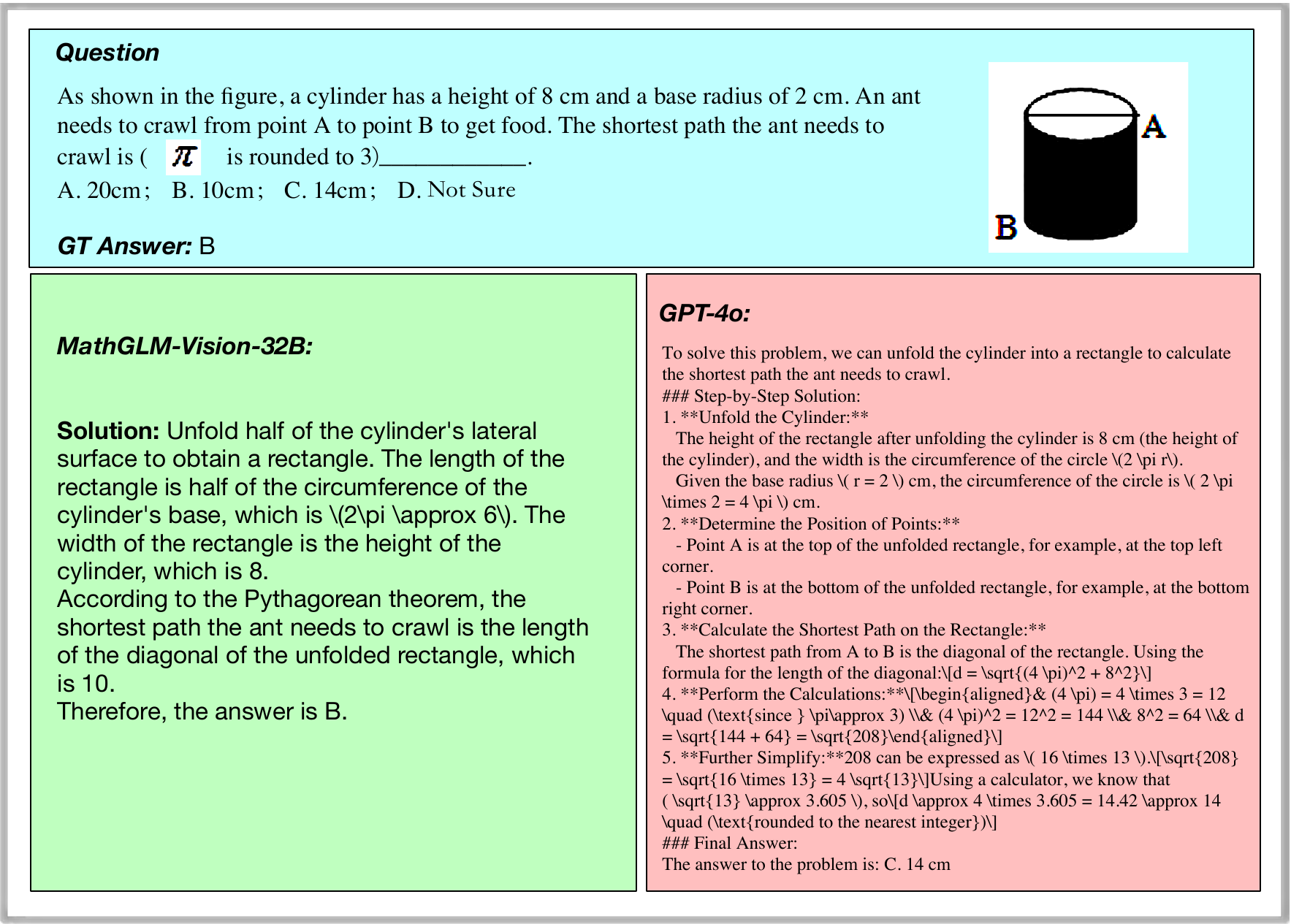}
    \vspace{-2mm}
    \caption{\textbf{An example of solid geometry problem. } \model-32B correctly calculated the shortest path with shorter steps, while GPT-4o made a reasoning error, leading to a different conclusion.}
    \label{fig:case_study_1}
    \vspace{-3mm}
\end{figure}

\begin{figure}[hbpt]
    \centering
    \includegraphics[width=1.0\textwidth]{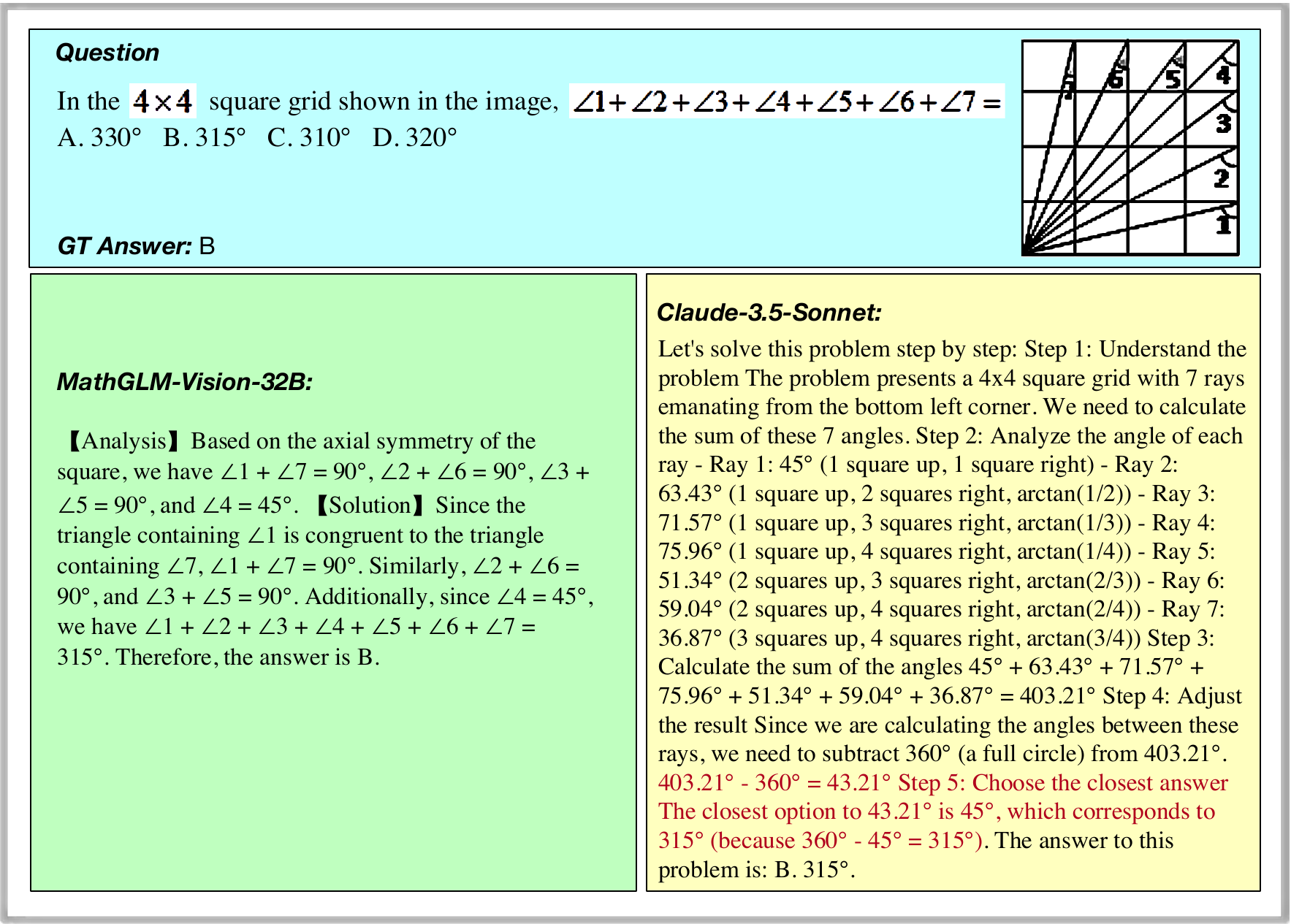}
    \vspace{-2mm}
    \caption{\textbf{An example of a planar geometry problem. } \model-32B correctly utilized axial symmetry to determine the angle sum, while Claude-3.5-Sonnet arrived at the correct answer through an erroneous calculation process. Despite the correct final answer, the calculation process was incorrect and overly complex. }
    \label{fig:case_study_2}
    \vspace{-3mm}
\end{figure}

\begin{figure}[hbpt]
    \centering
    \includegraphics[width=1.0\textwidth]{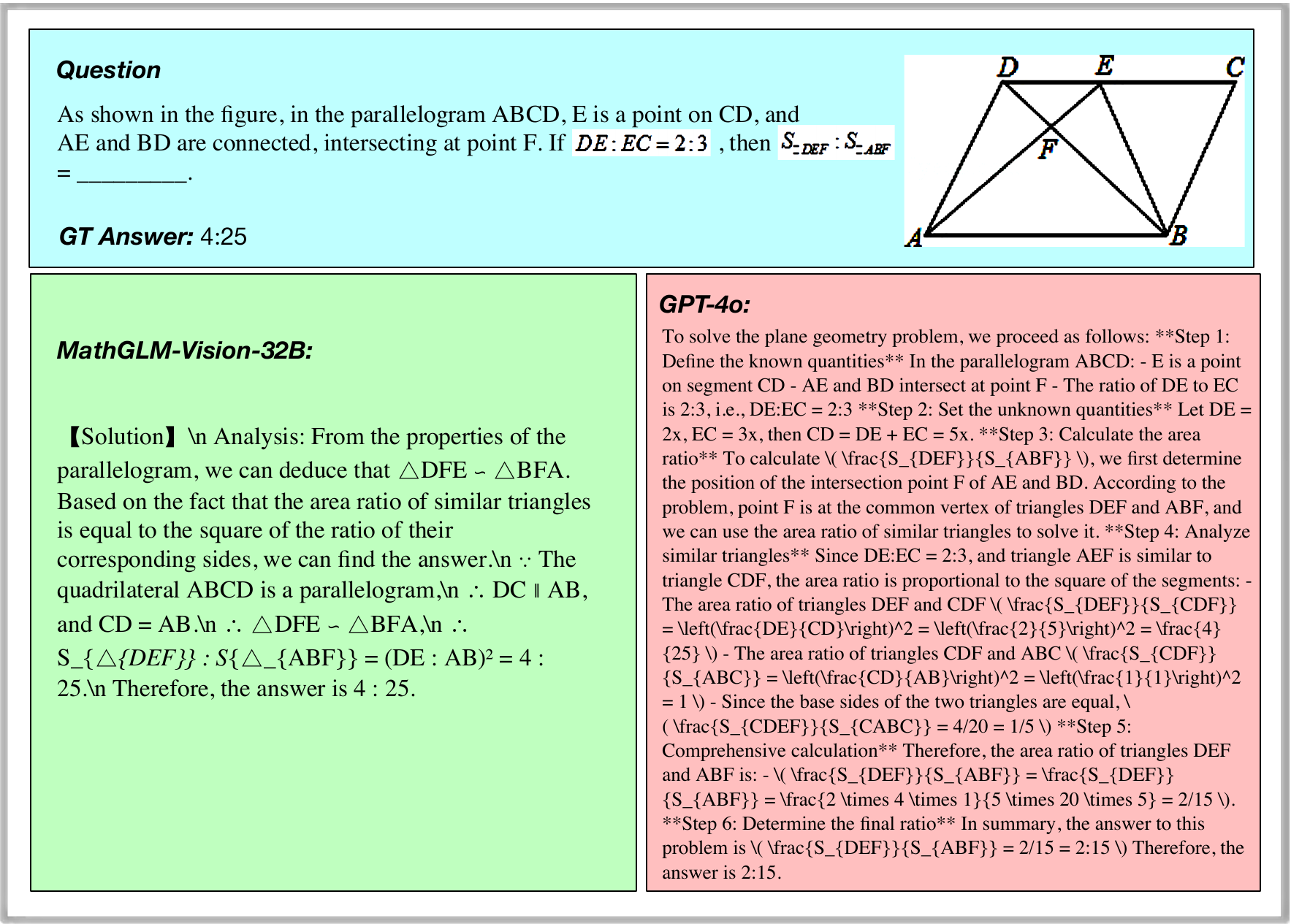}
    \vspace{-2mm}
    \caption{\textbf{An example of a planar geometry problem. } \model-32B correctly used the properties of similar triangles to find the area ratio, while GPT-4o misinterpreted the geometric relationships and misapplied the formula for the area ratio, leading to an erroneous conclusion.}
    \label{fig:case_study_3}
    \vspace{-3mm}
\end{figure}

\begin{figure}[hbpt]
    \centering
    \includegraphics[width=1.0\textwidth]{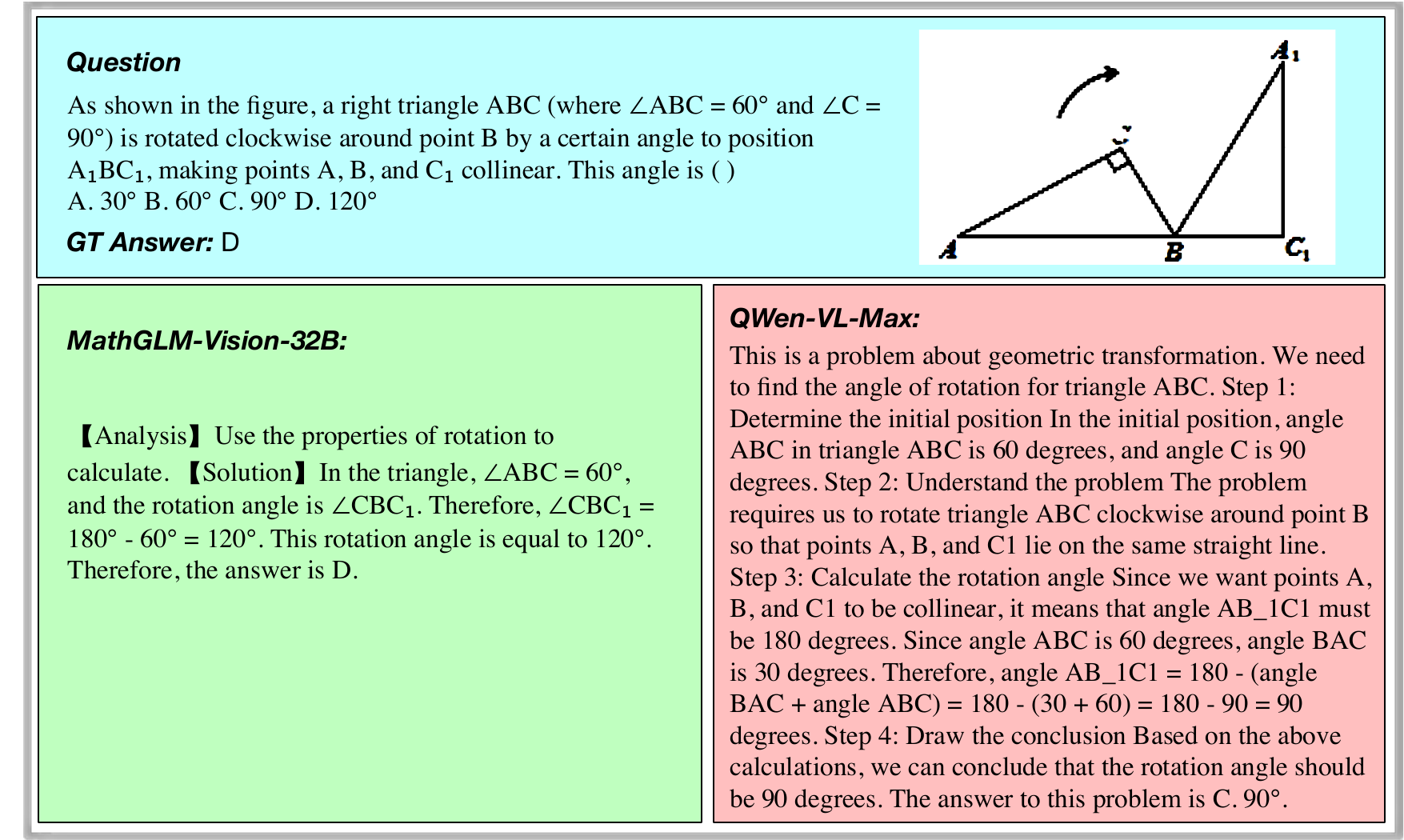}
    \vspace{-2mm}
    \caption{\textbf{An example of a geometric transformation problem. } \model-32B correctly used the properties of rotation to determine the angle, while QWen-VL-Max misunderstood the problem requirements and incorrectly calculated the rotation angle as 90°. }
    \label{fig:case_study_4}
    \vspace{-3mm}
\end{figure}

Figure ~\ref{fig:case_study_1}, Figure~\ref{fig:case_study_2}, Figure~\ref{fig:case_study_3}, and Figure~\ref{fig:case_study_4} present several case studies from \model-32B. These figures showcase the model's performance in various scenarios, highlighting its strengths in providing concise and clear answers with logical mathematical reasoning. Compared to other close-source MLLMs, \model-32B stands out for its ability to deliver precise and understandable solutions.

\section{Model Evaluation}

\vpara{Evaluation on public benchmarks}
The existing public benchmarks for evaluating a wide array of open-source and close-source models are neither timely nor comprehensive enough. To compare our \model with the state-of-the-art open-source and close-source LLMs, we have supplemented the evaluations for some models missing from the public benchmark leaderboard. 

We generate LLMs' responses through API access (for closed-source models) and local inference (for open-source models). The evaluation was then conducted following the official evaluation code from each benchmark's GitHub repository. The source of the models used in the evaluation can be found in Table ~\ref{tab:source_of_models}.

\begin{table*}[htbp]
    \centering
    \renewcommand{\arraystretch}{1.15}
    \small
    \resizebox{\textwidth}{!}{%
    \begin{tabular}{cccccccc}  
    \toprule
     Model & Input & LLM Size  & Source \\
    \midrule
    \multicolumn{4}{c}{\textit{Closed Source Models}}\\
     \midrule
     \multicolumn{4}{l}{\textit{Multi-modal LLMs}}\\
     Gemini Pro &    \textit{Q}, \textit{I} & - & \href{https://ai.google.dev/}{gemini-pro} \\
     Gemini 1.5 Pro &    \textit{Q}, \textit{I} & - & \href{https://ai.google.dev/}{gemini-1.5-pro} \\ 
     GPT-4V & \textit{Q}, \textit{I} & - & \href{https://platform.openai.com/}{gpt-4-vision-preview}  \\ 
     GPT-4-turbo & \textit{Q}, \textit{I} & - & \href{https://platform.openai.com/docs/models/gpt-4-turbo-and-gpt-4}{gpt-4-turbo} \\ 
     GPT-4o & \textit{Q}, \textit{I} & - & \href{https://platform.openai.com/docs/models/gpt-4o}{gpt-4o} \\ 
     Claude-3-Opus & \textit{Q}, \textit{I} & - &\href{https://www.anthropic.com/api}{claude-3-opus-20240229}\\
     Claude-3.5-Sonnet & \textit{Q}, \textit{I} & - &\href{https://www.anthropic.com/api}{claude-3-5-sonnet-2024620}\\
     Qwen-VL-Plus & \textit{Q}, \textit{I} & -  & \href{https://help.aliyun.com/zh/dashscope/developer-reference/vl-plus-quick-start}{qwen-vl-plus}\\
     Qwen-VL-Max & \textit{Q}, \textit{I} & - &  \href{https://help.aliyun.com/zh/dashscope/developer-reference/vl-plus-quick-start}{qwen-vl-max}\\
     \midrule
    \multicolumn{4}{c}{\textit{Open Source Models}}\\
    \midrule
    \multicolumn{4}{l}{\textit{General Multi-modal LLMs}}\\
    mPLUG-Owl  & \textit{Q}, \textit{I} & 7B & \href{https://github.com/X-PLUG/mPLUG-Owl}{mPLUG-Owl} &  \\
    LLaMA-Adapter-V2  & \textit{Q}, \textit{I} & 7B  & \href{https://github.com/ml-lab/LLaMA-Adapter-2}{LLaMA-Adapter V2} &    \\
    InstructBLIP  & \textit{Q}, \textit{I} & 7B & \href{https://huggingface.co/docs/transformers/main/en/model_doc/instructblip}{InstructBLIP} &   \\
    LLaVA-1.5  & \textit{Q}, \textit{I} & 13B & \href{https://github.com/haotian-liu/LLaVA}{LLaVA-v1.5-13B} &  \\
    ShareGPT-4V & \textit{Q}, \textit{I} & 13B & \href{https://huggingface.co/Lin-Chen/ShareGPT4V-13B}{ShareGPT4V-13B} &   \\
    SPHINX-MoE & \textit{Q}, \textit{I} & 8*7B & \href{https://github.com/Alpha-VLLM/LLaMA2-Accessory/blob/main/SPHINX/README.md}{SPHINX-MoE} & \\
    SPHINX-Plus & \textit{Q}, \textit{I} & 13B & \href{https://github.com/Alpha-VLLM/LLaMA2-Accessory/blob/main/SPHINX/README.md}{SPHINX-Plus} & \\
    InternLM-XC2 & \textit{Q}, \textit{I} & 7B & \href{https://huggingface.co/internlm/internlm-xcomposer2-vl-7b}{InternLM-XComposer2-VL-7B} & \\
    InternVL-1.2-Plus & \textit{Q}, \textit{I} & 34B & \href{https://huggingface.co/OpenGVLab/InternVL-Chat-V1-2-Plus}{InternVL-Chat-V1-2-Plus} & \\
    \midrule
    \multicolumn{4}{l}{\textit{Geo-Multi-modal LLMs}}\\
    G-LLaVA & \textit{Q}, \textit{I} & 7B & \href{https://huggingface.co/renjiepi/G-LLaVA-7B}{G-LLaVA-7B} &   \\
    G-LLaVA & \textit{Q}, \textit{I} & 13B & \href{https://huggingface.co/renjiepi/G-LLaVA-13B}{G-LLaVA-13B} &    \\
    LLaVA-1.5-G & \textit{Q}, \textit{I} & 7B & \href{https://huggingface.co/caishihao/GeoGPT4V-LLaVA-1.5-7B-v1}{LLaVA-1.5-7B-GeoGPT4V} &   \\
    LLaVA-1.5-G & \textit{Q}, \textit{I} & 13B & \href{https://huggingface.co/caishihao/GeoGPT4V-LLaVA-1.5-13B-v1}{LLaVA-1.5-13B-GeoGPT4V} & \\
    ShareGPT4V-G & \textit{Q}, \textit{I} & 7B & \href{https://huggingface.co/caishihao/GeoGPT4V-ShareGPT4V-7B-v1}{ShareGPT4V-7B-GeoGPT4V} & \\
    ShareGPT4V-G & \textit{Q}, \textit{I} & 13B & \href{https://huggingface.co/caishihao/GeoGPT4V-ShareGPT4V-13B-v1}{ShareGPT4V-1.5-13B-GeoGPT4V} & \\
    Math-LLaVA & \textit{Q}, \textit{I} & 13B & \href{https://github.com/HZQ950419/Math-LLaVA}{Math-LLaVA-13B} & \\
    \bottomrule
    \end{tabular}} 
    \caption{The source of the models used in the evaluation. }
    \label{tab:source_of_models}
\end{table*}

\vpara{Evaluation on \data-test}
We evaluate MathGLM-Vision and several close-source
MLLMs using our specially constructed \data-test. The evaluation process of our \data-test is conducted through 3 key-step: generation, extraction, and scoring. 

For the generation step, the model responses are generated by providing the model with queries which incorporate the Chain of Thought (CoT) template, questions, and diagram information. The reponses of close-source MLLMs is generated through API access. For the extraction step, we use GPT-3.5-turbo to extract the model's answer based on the reponses of first step. Finally, in the scoring step, the score for each question is determined by GLM-4 based on the comparison between the extracted answer and the standard answer.

The prompts used to guide the LLM in response generation, answer extraction and scoring can be found in Table ~\ref{tab:prompts}.

\begin{table}[hbpt]
    \centering
    \renewcommand{\arraystretch}{1.15}
    \resizebox{\textwidth}{!}{%
    \begin{tabular}{>{\centering\arraybackslash}m{3cm}|m{10cm}}
    \toprule
     Task    & Prompt  \\
    \midrule
    \centering Response Generation & You are a very skilled math teacher. Please provide a detailed, step-by-step solution to the question, following a step-by-step format. Be sure to conclude with a summary that states "The answer to this question is" followed by the final result.\\
    \hline
     \centering Answer Extraction & Please read the following example. Then extract the answer from the model response and type it at the end of the prompt. Hint: Please answer the question requiring an integer answer and provide the final value, e.g., 1, 2, 3, at the end. Question: Which number is missing? Model response: The number missing in the sequence is 14. Extracted answer: 14 Hint: Please answer the question requiring a floating-point number with one decimal place and provide the final value, e.g., 1.2, 1.3, 1.4, at the end. Question: What is the fraction of females facing the camera? Model response: The fraction of females facing the camera is 0.6, which means that six out of ten females in the group are facing the camera. Extracted answer: 0.6\\
    \hline     
     \centering Scoring &  Please determine if the extracted\_answer correctly answers the question. The correct answer needs to be extracted from the answer without recalculating it, and the answer in the answer should be considered the final answer. Also, do not judge whether the answer is correct. The question may contain multiple sub-questions, and correctly answering the question includes correctly answering every sub-question and every result within each sub-question. A relative error divided by the absolute value of the original answer of less than 0.01 is allowed. If the prediction does not contain an answer, it is considered wrong. If the answer is not numerical, determine the equivalence of the expression, not just the value. If there is one mistake, the answer is wrong. Only if all results given in the prediction are correct is it considered correct. There is no need to consider whether the solution process of the prediction is complete. Please first extract the answers given by the prediction, determine the relative error, check if each sub-question is answered correctly, and finally give the judgment in a single line (output only "yes" or "no" in a single line).  \\
    \bottomrule
    \end{tabular}} 
    \vspace{0.2cm}
    \caption{Prompts for response generation, answer extraction and scoring.}
    \label{tab:prompts}
\end{table}

\end{document}